\documentclass{article}

\usepackage{microtype}
\usepackage{graphicx}
\usepackage{subfigure}
\usepackage{booktabs} 

\usepackage{amsmath,amscd,amssymb}
\usepackage{array}
\usepackage{bbold}
\usepackage{bm}
\usepackage{mathtools}

\newcommand{\bth}{{\bm{\theta}}}
\newcommand{\beps}{{\bm{\epsilon}}}
\newcommand{\bz}{{\bm{z}}}
\newcommand{\bx}{{\bm{x}}}
\usepackage{enumitem}

\usepackage[
    acronym,smallcaps
]{glossaries} 
\newacronym[\glslongpluralkey={deep exponential families}]{DEF}{def}{deep exponential family}
\newacronym[\glslongpluralkey={rsvi}]{RSVI}{rsvi}{rsvi}
\newacronym[\glslongpluralkey={G-Rep}]{G-Rep}{G-Rep}{generalized reparameterization gradient}

\newcommand{\E}{\mathbb{E}}

\usepackage{hyperref}

\usepackage[accepted]{icml2018}

\icmltitlerunning{Pathwise Derivatives Beyond the Reparameterization Trick}

\begin{document}

\twocolumn[
\icmltitle{Pathwise Derivatives Beyond the Reparameterization Trick}



\icmlsetsymbol{equal}{*}

\begin{icmlauthorlist}
\icmlauthor{Martin Jankowiak}{equal,uail}
\icmlauthor{Fritz Obermeyer}{equal,uail}
\end{icmlauthorlist}

\icmlaffiliation{uail}{Uber AI Labs, San Francisco, USA}

\icmlcorrespondingauthor{}{jankowiak@uber.com}
\icmlcorrespondingauthor{}{fritzo@uber.com}

\icmlkeywords{Machine Learning, ICML}

\vskip 0.3in
]



\printAffiliationsAndNotice{\icmlEqualContribution} 

\begin{abstract}
We observe that gradients computed via the reparameterization trick are in direct correspondence with solutions of the transport equation in the formalism of optimal transport. We use this perspective to compute (approximate) pathwise gradients for probability distributions not directly amenable to the reparameterization trick: Gamma, Beta, and Dirichlet. We further observe that when the reparameterization trick is applied to the Cholesky-factorized multivariate Normal distribution, the resulting gradients are suboptimal in the sense of optimal transport. 
We derive the optimal gradients and show that they have reduced variance in a Gaussian Process regression task. We demonstrate
with a variety of synthetic experiments and stochastic variational inference tasks that our pathwise gradients are competitive with other methods.
\end{abstract}

\section{Introduction}

Maximizing objective functions via gradient methods is ubiquitous in machine learning. When the objective function $\mathcal{L}$
is defined as an expectation of a (differentiable) test function $f_\bth(\bz)$ w.r.t.~a probability distribution $q_{\bth}(\bz)$,
\begin{equation}
\mathcal{L} = \E_{q_{\bth}(\bz)} \left[ f_\bth(\bz) \right]
\end{equation}
computing exact gradients w.r.t.~the parameters $\bth$ is often unfeasible so that optimization
methods must instead make due with stochastic gradient estimates. If the gradient estimator is unbiased, then stochastic gradient
descent with an appropriately chosen sequence of step sizes can be shown to have nice convergence properties \cite{robbins1951stochastic}. 
If, however, the gradient estimator exhibits large variance, stochastic optimization algorithms may be impractically slow. Thus
it is of general interest to develop gradient estimators with reduced variance. 

We revisit the class of gradient estimators popularized in \cite{kingma2013auto,rezende2014stochastic,titsias2014doubly}, which go under the name of the pathwise derivative or the reparameterization trick. While this class of gradient estimators is not applicable to all
choices of probability distribution $q_{\bth}(\bz)$, empirically it has been shown to yield suitably low variance in many cases of
practical interest and thus has seen wide use. We show that the pathwise derivative in the literature is in fact a 
particular instance of a continuous family of gradient estimators. Drawing a connection to tangent fields in the field of optimal 
transport,\footnote{See \cite{villani2003topics,ambrosio2008gradient} for a review.} 
we show that one can define a unique pathwise gradient that is optimal in the sense of optimal transport. For the purposes
of this paper, we will refer to these optimal gradients as OMT (optimal mass transport)  gradients.

\begin{figure}[t]
\begin{center}
\centerline{\includegraphics[width=.9\columnwidth]{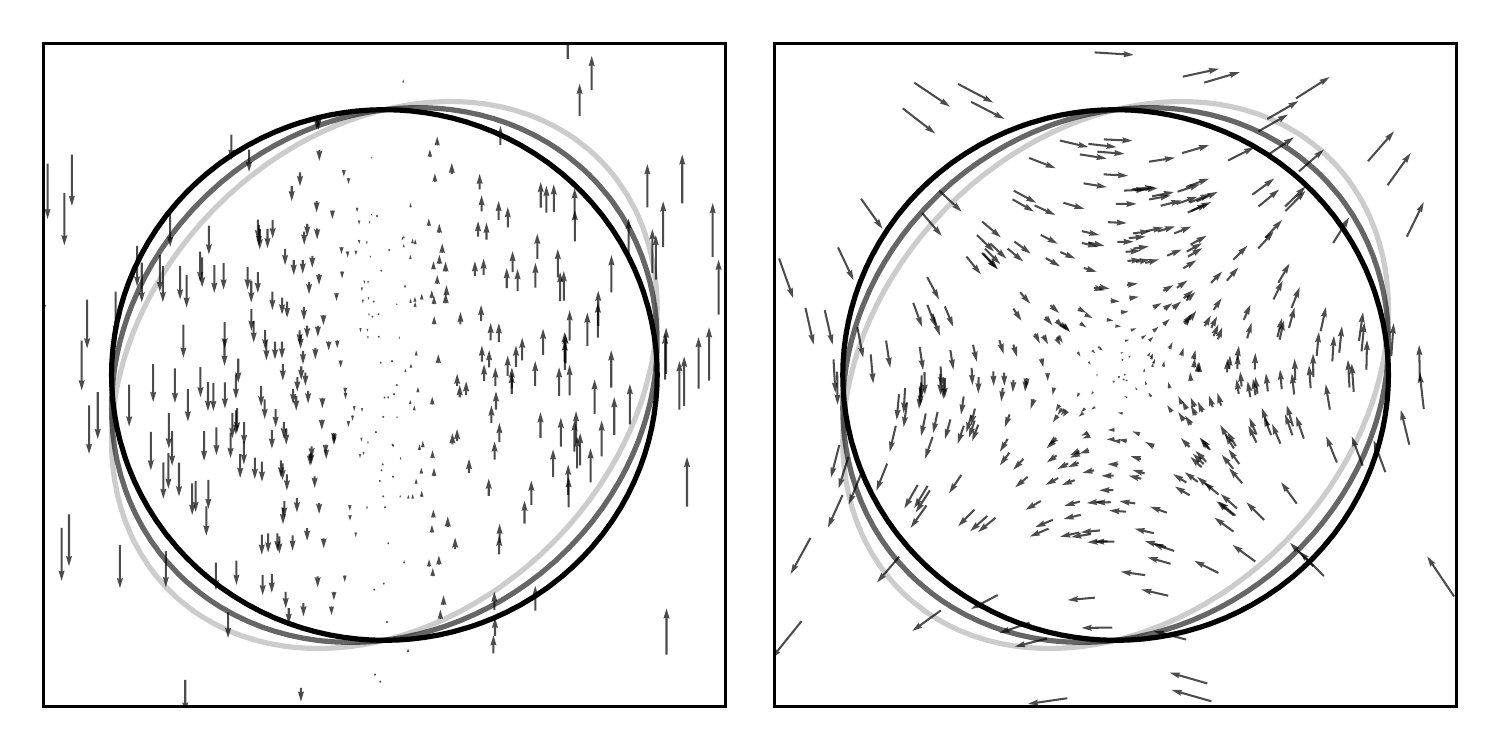}}
\caption{Velocity fields for a bivariate Normal distribution parameterized by a Cholesky factor $\bm{L}=\mathbb{1}_2$. 
The gradient is w.r.t.~the off-diagonal element $L_{21}$. On the left we depict the velocity field corresponding
to the reparameterization trick and on the right we depict the velocity field that is optimal in the sense of optimal
transport. The solid black circle denotes the 1-$\sigma$ covariance ellipse, with the gray ellipses denoting displaced
 covariance ellipses that result from small increases in $L_{21}$. Note that the ellipses evolve the same way under both velocity fields, but
 \emph{individual} particles flow differently to effect the same global displacement of mass.}
\label{fig:mvn}
\end{center}
\end{figure}

The resulting geometric picture is particularly intriguing in the case of multivariate distributions, where each choice of gradient estimator specifies a velocity field on the sample space. 
To make this picture more concrete, in Figure \ref{fig:mvn} we show the velocity fields that correspond to two different gradient estimators for the off-diagonal element of the Cholesky factor parameterizing a bivariate Normal distribution. We note that the velocity field that corresponds to the reparameterization trick has a large rotational component that makes it suboptimal in the sense of optimal transport.
In Sec.~\ref{sec:exp} we show that this suboptimality can result in reduced performance when fitting a Gaussian Process to data.

The rest of this paper is organized as follows.
In Sec.~\ref{sec:svi} we provide a brief overview of stochastic gradient variational inference.
In Sec.~\ref{sec:crg} we show how to compute pathwise gradients for univariate distributions.
In Sec.~\ref{sec:multi} we expand our discussion of pathwise gradients to the case of multivariate distributions, introduce
the connection to the transport equation,
and provide an analytic formula for the OMT gradient in the case of the multivariate Normal.
In Sec.~\ref{sec:numerical} we discuss how we can compute high precision approximate pathwise gradients for the Gamma, Beta, and Dirichlet distributions. 
In Sec.~\ref{sec:relatedwork} we place our work in the context of related research.
In Sec.~\ref{sec:exp} we demonstrate the performance of our gradient estimators with a variety of synthetic experiments and
experiments on real world datasets.
Finally, in Sec.~\ref{sec:discussion} we conclude with a discussion of directions for future work.

\vspace{-3pt}

\section{Stochastic Gradient Variational Inference}
\label{sec:svi}

One area where stochastic gradient estimators play a particularly central role is stochastic variational inference \cite{hoffman2013stochastic}.
This is especially the case for black-box methods \cite{wingate2013automated,ranganath2014black}, where conjugacy and other simplifying structural assumptions are unavailable, with the consequence that Monte Carlo estimators become necessary. For concreteness, we will refer to this class
of methods as Stochastic Gradient Variational Inference (SGVI).  In this section we give a brief overview of this line of research, as it serves as the motivating use case for our work. Furthermore, in Sec.~\ref{sec:exp} SGVI will serve as the main testbed for our proposed methods.

Let $p(\bx, \bz)$ define a joint probability distribution over observed data $\bx$ and latent random variables $\bz$. One of the main
tasks in Bayesian inference is to compute the posterior distribution $p(\bz|\bx) = \frac{p(\bx,\bz)}{p(\bx)}$. For many models of interest,
this is an intractably hard problem and so approximate methods become necessary. Variational inference recasts Bayesian
inference as an optimization problem. Specifically we define a variational family of distributions $q_\bth(\bz)$ parameterized by 
$\bth$ and seek to find a value of $\bth$ that minimizes the KL divergence between $q_\bth(\bz)$ and the (unknown) posterior $p(\bz|\bx)$. This is equivalent to maximizing the ELBO \cite{jordan1999introduction}, defined as
\begin{equation}
\begin{split}
{\rm ELBO} = \E_{q_\bth(\bz)} \left[ \log p(\bx,\bz) - \log q_\bth(\bz) \right]
\end{split}\label{eq:elbo}
\end{equation}
For general choices of $p(\bx,\bz)$ and $q_\bth(\bz)$, this expectation---much less its gradients---cannot be computed analytically. In these
circumstances a natural approach is to build a Monte Carlo estimator of the ELBO and its gradient w.r.t.~$\bth$.
The properties of the chosen gradient estimator---especially its bias and variance---play a critical rule in determining the viability of the resulting stochastic optimization. Next, we review two commonly used gradient estimators; we leave a brief discussion of more elaborate variants to Sec.~\ref{sec:relatedwork}. 

\subsection{Score Function Estimator}

The score function estimator, also referred to as the log-derivative trick or \textsc{reinforce} \cite{glynn1990likelihood,williams1992simple}, provides a simple and broadly applicable recipe for estimating ELBO 
gradients \cite{paisley2012variational}. 
The score function estimator expresses the gradient as an expectation with respect to $q_\bth(\bz)$, with the simplest
variant given by 
\begin{equation}
\label{eqn:scorefunction}
\nabla_\bth {\rm ELBO} =  \E_{q_{\bth}(\bz)} \left[ \nabla_\bth \log r + \log r \nabla_\bth \log q_{\bth}(\bz) \right]
\end{equation}
where $\log r = \log p(\bx,\bz) - \log q_{\bth}(\bz)$. Monte Carlo estimates of Eqn.~\ref{eqn:scorefunction} can 
be formed by drawing samples from $q_\bth(\bz)$ and computing the term in the square brackets. 
Although the score function estimator is very general (e.g.~it applies to discrete random variables) it
typically suffers from high variance, although this can be mitigated with the use of variance reduction techniques
such as Rao-Blackwellization \cite{casella1996rao} and control variates \cite{ross}.

\subsection{Pathwise Gradient Estimator}

The pathwise gradient estimator, a.k.a.~the reparameterization trick (RT), is not as broadly applicable
as the score function estimator, but it generally exhibits lower variance 
\cite{price1958useful,salimans2013fixed,kingma2013auto,glasserman2013monte,rezende2014stochastic,titsias2014doubly}. 
It is applicable to continuous
random variables whose probability density $q_{\bth}(\bz)$ can be reparameterized such that we can rewrite expectations
\begin{equation}
\label{eqn:reptrick}
\E_{q_{\bth}(\bz)} \left[ f_\bth(\bz) \right] \;\; \longrightarrow \;\;
\E_{q_0(\beps)} \left[ f_\bth(\mathcal{T}(\beps; \bth)) \right ] 
\end{equation}
where $q_0(\bz)$ is a fixed distribution with no dependence on $\bth$ and $\mathcal{T}(\beps; \bth)$ is a differentiable
$\bth$-dependent transformation. 
Since the expectation w.r.t.~$q_0(\beps)$ has no $\bth$ dependence, gradients w.r.t.~$\bth$
can be computed by pushing $\nabla_\bth$ through the expectation. 
This reparameterization can be done for a number of distributions, including for example the Normal distribution.
Unfortunately the reparameterization trick is non-trivial to apply to a number of commonly used distributions, e.g.~the Gamma and Beta distributions, since the required shape transformations $\mathcal{T}(\beps; \bth)$ inevitably involve special functions.

\section{Univariate Pathwise Gradients}
\label{sec:crg}

Consider an objective function given as the expectation of a test function $f_\bth(z)$ with respect
to a distribution $q_\bth(z)$, where $z$ is a continuous one-dimensional random variable:
\begin{equation}
\mathcal{L} = \E_{q_{\bth}(z)} \left[ f_\bth(z) \right]
\end{equation}
Here $q_\bth(z)$ and $f_\bth(z)$ are parameterized by $\bth$, and 
we would like to compute (stochastic) gradients of $\mathcal{L}$ w.r.t.~$\theta$,
where $\theta$ is a scalar component of $\bth$:
\begin{equation}
\label{eqn:targetgrad}
\nabla_\theta \mathcal{L} = \nabla_\theta \E_{q_{\bth}(z)} \left[ f_\bth(z) \right]
\end{equation}
Crucially we would like to avoid the log-derivative trick, which yields a gradient estimator that tends to have high variance.
Doing so will be easy if we can rewrite the expectation in terms of a fixed
distribution that does not depend on $\theta$. A natural choice is to use the standard uniform distribution $\mathcal{U}$,
\begin{equation}
\mathcal{L} = \E_{\mathcal{U}(u)} \left[ f_\bth(F_\bth^{-1}(u) ) \right]
\end{equation}
where the transformation $F_\bth^{-1}: u \rightarrow z$ is the inverse CDF of $q_\bth(z)$. 
As desired, all dependence on $\theta$ is now inside the expectation. Unfortunately, for many continuous univariate distributions
of interest (e.g.~the Gamma and Beta distributions) the transformation $F_\bth^{-1}$ (as well as its derivative w.r.t.~$\bth$) does not admit a simple analytic expression. 

Fortunately, by making use of implicit differentiation we can compute the gradient in Eqn.~\ref{eqn:targetgrad} without explicitly introducing $F_\bth^{-1}$. To complete the derivation define $u$ by  
\begin{equation}
\label{eqn:udef}
u \equiv F_\bth(z) = \int_{-\infty}^z q_\bth(z^\prime) dz^\prime
\end{equation}
and  differentiate both sides of Eqn.~\ref{eqn:udef} w.r.t.~$\theta$ and make use of the fact that $u\sim \mathcal{U}$ does not depend on $\theta$ to obtain
\begin{equation}
0 = \frac{dz}{d\theta} q_\bth(z) + \int_{-\infty}^z \frac{\partial}{\partial \theta}q_\bth(z^\prime) dz^\prime
\end{equation}
This then yields our master formula for the univariate case
\begin{equation}
\label{eqn:master}
\frac{dz}{d\theta}  = -\frac{\frac{\partial F_\bth}{\partial \theta }(z)}{q_\bth(z)}
\end{equation}
where the corresponding gradient estimator is given by
\begin{equation}
\label{eqn:1destimator}
\nabla_\theta \mathcal{L} = \E_{q_{\bth}(z)} \left[ \frac{ d f_\bth(z)}{d z} \frac{dz}{d\theta}  + \frac{\partial f_\bth(z)}{\partial \theta}\right]
\end{equation}
While this derivation is elementary, it helps to clarify things:
the key ingredient needed to compute pathwise gradients in Eqn.~\ref{eqn:targetgrad} 
is the ability to compute (or approximate) the derivative of the CDF, i.e.~$\frac{\partial}{\partial \theta} F_{\bth}(z)$. 
In the supplementary materials we verify that Eqn.~\ref{eqn:1destimator} results in correct gradients.

It is worth emphasizing how this approach differs from a closely related alternative. 
Suppose we construct a (differentiable) approximation of the \emph{inverse} CDF, $\hat{F}_{\bth}^{-1}(u) \approx F_{\bth}^{-1}(u)$. 
For example, we might train a neural network ${\rm nn}(u, \bth) \approx F_{\bth}^{-1}(u)$. We can then push samples
$u \sim \mathcal{U}$ through ${\rm nn}(u, \bth)$ and obtain approximate samples from $q_{\bth}(z)$ as well as approximate
derivatives $\frac{dz}{d\theta}$ via the chain rule;
in this case, there will be a mismatch between the probability $q_\bth(z)$ assigned to samples $z$ and the
actual distribution over $z$. By contrast, if we use the construction of Eqn.~\ref{eqn:master}, our samples $z$ will still be
exact\footnote{Or rather their exactness will be determined by the quality of our sampler for $q_\bth(z)$, which is fully decoupled from how
we compute derivatives $\frac{dz}{d\theta}$.} and the fidelity of our approximation of (the derivatives of) $F_{\bth}(z)$ will only affect the accuracy of our 
approximation for $\frac{dz}{d\theta}$.

\section{Multivariate Pathwise Gradients}
\label{sec:multi}

In the previous section we  focused on continuous univariate distributions. Pathwise gradients can also be constructed 
for continuous multivariate distributions, although the analysis is in general expected to be 
much more complicated than in the univariate 
case---directly analogous to the difference between ordinary and partial differential equations. Before constructing estimators for particular distributions,
we introduce the connection to the transport equation.

\subsection{The Transport Equation}

Consider a multivariate distribution $q_\bth(\bm{z})$ in $D$ dimensions and
consider differentiating $\E_{q_{\bth}(\bm{z})} \left[ f(\bm{z}) \right]$ with respect to the parameter $\theta$.\footnote{Here 
without loss of generality we assume that $f(\bm{z})$ has no dependence on $\theta$, since computing 
$\E_{q_{\bth}(\bm{z})} \left[ \nabla_{\theta} f_\bth(\bm{z}) \right]$ presents no difficulty; the difficulty stems from 
the dependence on $\theta$ in $q_\bth(z)$. }  As we vary $\theta$ we move
$q_{\bth}(\bm{z})$ along a curve in the space of distributions over the sample space. Alternatively, we can think of each
distribution as a cloud of particles; as we vary $\theta$  from $\theta$ to 
$\theta +\Delta \theta$ each particle undergoes an 
infinitesimal displacement $d\bm{z}$. Any set of displacements that ensures that the displaced particles are
distributed according to the displaced distribution $q_{\theta + \Delta \theta}(\bm{z})$ is allowed. 
This intuitive picture can be formalized with
the transport a.k.a.~continuity equation:\footnote{We refer the reader to \cite{villani2003topics} and \cite{ambrosio2008gradient} for details.}
\begin{equation}
\label{eqn:transport}
\frac{\partial}{\partial \theta} q_\bth + \nabla_{\bm z} \cdot \left( q_\bth {\bm v^\theta}\right)=0
\end{equation}
Here the \emph{velocity field} $\bm{v}^\theta$ is a vector field defined on the sample space that 
displaces samples (i.e.~particles) $\bm z$ as we vary $\theta$ infinitesimally. Note that there
is a velocity field $\bm{v}^\theta$ for each component $\theta$ of $\bth$.
This equation is readily interpreted in the language of fluid dynamics.
In order for the the total probability to be conserved, the term $\frac{\partial}{\partial \theta} q_\bth({\bm z})$---which is the rate of change
of the number of particles in the infinitesimal volume element at $\bm z$---has to be counterbalanced by the in/out-flow of particles---as given by the divergence 
term.

\subsection{Gradient Estimator}

Given a solution to Eqn.~\ref{eqn:transport}, we can form the gradient estimator
\begin{equation}
\label{eqn:multiest}
\nabla_\theta \mathcal{L} = \E_{ q_\bth(\bz) } \! \left[ \bm{v}^\theta  \cdot \nabla_{\bf z} f  \right]
\end{equation}
which generalizes Eqn.~\ref{eqn:1destimator} to the multivariate case. That this is an
unbiased gradient estimator follows directly from the divergence theorem (see the supplementary materials).

\subsection{Tangent Fields}

In general Eqn.~\ref{eqn:transport} admits an infinite dimensional space of solutions. In the context of our derivation of Eqn.~\ref{eqn:master},
 we might loosely say that different solutions of Eqn.~\ref{eqn:transport} correspond to different ways of specifying
 quantiles of $q_{\bth}(\bm{z})$. To determine a \emph{unique}\footnote{We refer the reader to Ch.~8 of \cite{ambrosio2008gradient} for details.} solution---the tangent field from the theory of optimal transport---we require that
\begin{equation}
\label{eqn:symm}
\frac{\partial v_i^{\rm OMT}}{\partial z_j} = \frac{\partial v_j^{\rm OMT}}{\partial z_i}  \qquad \forall i, j
\end{equation}
In this case it can be shown that $\bm{v}^{\rm OMT}$ minimizes the 
total kinetic energy, which is given by\footnote{Note that the univariate solution, Eqn.~\ref{eqn:master}, is automatically the OMT solution.}
\begin{equation}
\label{eqn:kin}
K(\bm{v}) = \tfrac{1}{2} \int \! d\bm{z}\; q_\bth(z) || \bm{v}||^2
\end{equation}

\subsection{Gradient variance}

The $||\bm{v}||^2$ term that appears in Eqn.~\ref{eqn:kin} might lead one to hope that
$\bm{v}^{\rm OMT}$  provides gradients that minimize gradient variance.
Unfortunately, the situation is more complicated. 
Denoting the (mean) gradient by $\bm{g} = \E_{ q_\bth(\bz) } \! \left[ \bm{v}  \cdot \nabla_{\bf z} f(\bz)  \right]$
the total gradient variance is given by
\begin{equation}
\label{eqn:variance}
\E_{ q_\bth(\bz) } \left[ ||\bm{v}  \cdot \nabla_{\bf z}f ||^2  \right] - ||\bm{g}||^2
\end{equation}
Since $\bm{g}$ is the same for all unbiased gradient estimators, the gradient estimator that minimizes
the total variance is the one that minimizes the first term in Eqn.~\ref{eqn:variance}.  
For test functions $f(\bm{z})$ that approximately satisfy $\nabla_{\bf z}f \propto \mathbb{1}$ over the bulk of the support
of $q_\bth(\bm{z})$, the first term in Eqn.~\ref{eqn:variance} term is approximately proportional to the kinetic energy. In this
case the OMT gradient estimator will be (nearly) optimal.  Note
that  the kinetic energy weighs contributions from different components of $\bm{v}$ equally, whereas $\bm{g}$
scales different components of $\bm{v}$ with $\nabla_{\bf \bm{z}}f$. Thus we can think
of the OMT gradient estimator as a good choice for generic choices of $f(\bm{z})$ that are relatively flat and isotropic (or, alternatively, for choices
of $f(\bm{z})$ where we have little \emph{a priori} knowledge about the detailed structure of $\nabla_{\bf \bm{z}}f$).
So for any particular choice of a generic $f(\bm{z})$ there will be some gradient estimator that has lower variance than the OMT gradient estimator. Still,
for \emph{many} choices of $f(\bm{z})$ we expect the OMT gradient estimator to have lower variance than the 
RT gradient estimator, since the latter has no particular optimality guarantees (at least not in any coordinate
system that we expect to be well adapted to $f(\bm{z})$).

\subsection{The Multivariate Normal}

In the case of a (zero mean) multivariate Normal distribution 
parameterized by a Cholesky factor $\bm{L}$ via $\bm{z} = \bm{L}\bm{\tilde{z}}$, where
$\bm{\tilde{z}}$ is white noise, the reparameterization trick yields the following velocity field for $L_{ab}$:\footnote{Note that
the reparameterization trick already yields the OMT gradient for the location parameter $\bm{\mu}$.}
\begin{equation}
\label{eqn:rtvec}
v_i^{\rm RT} = \frac{\partial z_i}{\partial L_{ab}} =  \delta_{ia}  (L^{-1} \bm{z})_b 
\end{equation}
Note that Eqn.~\ref{eqn:rtvec}  is just a particular instance of the solution to the transport
equation that is implicitly provided by the reparameterization trick, namely
\begin{equation}
{\bm v}^\theta = \frac{\partial \mathcal{T}(\beps; \bth)}{\partial \theta}\bigg\rvert_{\beps=\mathcal{T}^{-1}(\bz; \bth)}
\end{equation}
In the supplementary materials we verify that Eqn.~\ref{eqn:rtvec} satisfies the transport equation Eqn.~\ref{eqn:transport}.
However, it is evidently \emph{not} optimal in the sense of optimal transport, since
$\frac{\partial v_i^{\rm RT}}{\partial z_j} =  \delta_{ia}  L^{-1}_{bj} $
is not symmetric in $i$ and $j$.
In fact the tangent field takes the form
\begin{equation}
\label{eqn:mvnopt}
v_i^{\rm OMT} =  
\frac{1}{2} \left(  \delta_{ia}  (L^{-1} \bm{z})_b + z_a L^{-1}_{bi}   \right) + (S^{ab} \bm{z})_i
\end{equation}
where $S^{ab}$ is a symmetric matrix whose precise form we give in the supplementary materials. We
note that computing gradients with Eqn.~\ref{eqn:mvnopt}
is $\mathcal{O}(D^3)$, since it involves a singular value decomposition of the covariance matrix.
In Sec.~\ref{sec:exp} we show that the resulting gradient estimator can lead to reduced variance.


\section{Numerical Recipes}
\label{sec:numerical}

In this section we show how Eqn.~\ref{eqn:master} can be used to obtain pathwise gradients in practice. In many cases of interest we will need to derive approximations to $\frac{\partial}{\partial \theta} F(z)$ that balance the need for high accuracy (thus yielding gradient estimates with negligible bias) with the need for computational efficiency. In particular we will derive accurate approximations to Eqn.~\ref{eqn:master} for the Gamma, Beta, and Dirichlet distributions. These approximations will involve three basic components:
\begin{enumerate}[topsep=0pt,itemsep=0ex,partopsep=1ex,parsep=1ex]
\item Elementary Taylor expansions
\item The Lugannani-Rice saddlepoint expansion \cite{lugannani1980saddle, butler2007saddlepoint}
\item Rational polynomial approximations in regions of $(z, \theta)$ that are analytically intractable
\end{enumerate}

\subsection{Gamma}
\label{sec:gamma}

The CDF of the Gamma distribution involves the (lower) incomplete gamma function $\gamma(\cdot)$:
$F_{\alpha, \beta}(z) = \frac{\gamma(\alpha, \beta z)}{\Gamma(\alpha)}$.
Unfortunately $\gamma(\cdot)$ does not admit simple analytic expressions for derivatives w.r.t.~its first argument, and so we must resort to numerical approximations.
Since $z \sim \rm{Gamma}(\alpha, \beta=1) \Leftrightarrow z/\beta  \sim \rm{Gamma}(\alpha, \beta)$  it is sufficient to consider 
$\frac{dz}{d\alpha}$ for the standard Gamma distribution with $\beta=1$.

\subsubsection{$z \ll 1$}
To give a flavor for the kinds of approximations we use, consider how we can approximate 
$\frac{\partial}{\partial \alpha} \gamma(\alpha, z)$ in the limit $z \ll 1$.
We simply do a Taylor series in powers of $z$: 
\begin{equation}
\begin{split}
\nonumber
\frac{\partial}{\partial \alpha} \gamma(\alpha, z) & = 
\frac{\partial}{\partial \alpha} \int_{0}^z (z^\prime)^\alpha \left(1/z^\prime - 1 + \tfrac{1}{2}z^\prime + ... \right) dz^\prime \\
 & = \frac{\partial}{\partial \alpha} z^\alpha \left(\frac{1}{\alpha} -\frac{z}{\alpha+1} + \frac{ \tfrac{1}{2}z^2}{\alpha+2} + ... \right)
\end{split}
\end{equation}
In practice we use 6 terms in this expansion, which is accurate
for $z < 0.8$.
 Details for the remaining approximations 
 can be found in the supplementary materials.

\subsection{Beta}
\label{sec:beta}

The CDF of the Beta distribution, $F_{\rm Beta}$, is the (regularized) incomplete beta function; 
just like in the case of the Gamma distribution, its derivatives do not admit simple analytic expressions. 
We describe the numerical approximations we used in the supplementary materials.

\subsection{Dirichlet}
\label{sec:dirichlet}

Let $\bm{z} \sim \rm{Dir}(\bm{\alpha})$ be Dirichlet distributed with $n$ components. Noting that
the $z_i$ are constrained to lie within the unit $(n-1)$-simplex, we proceed by representing
$\bm{z}$ in terms of $n-1$ mutually independent Beta variates \cite{wilks}:
\begin{equation}
\nonumber
\label{eqn:dirichletfrombeta}
\begin{split}
&\tilde{z}_i \sim \rm{Beta}(\alpha_i, {\scriptstyle \sum}_{j=i+1}^n \alpha_j) \qquad \rm{for} \qquad i=1,...,n-1 \\
& z_1 = \tilde{z}_1 \qquad\qquad  z_n = {\scriptstyle\prod}_{j=1}^{n-1} (1-\tilde{z}_j ) \\ 
& z_i = \tilde{z}_i {\scriptstyle \prod}_{j=1}^{i-1} (1-\tilde{z}_j ) \qquad \rm{for} \qquad i=2,...,n-1 \\
\end{split}
\end{equation}
Without loss of generality, we will compute $\frac{d}{d\alpha_1} z_i$ for $i=1,...,n$. 
Crucially, the only dependence on $\alpha_1$ in Eqn.~\ref{eqn:dirichletfrombeta} is through $\tilde{z}_1$.
We find:
\begin{equation}
\label{eqn:dirichletmaster}
\frac{d\bm{z}}{d\alpha_1} = -\frac{\frac{\partial F_{\rm{Beta}}}{\partial \alpha_1 }(z_1 | \alpha_1, \alpha_{\rm tot} - \alpha_1)}{\rm{Beta}(z_1 | \alpha_1, \alpha_{\rm tot} - \alpha_1)} \times \left( 1, \frac{-z_2}{1-z_1}, ..., \frac{-z_n}{1-z_1} \right)
\end{equation}
Note that Eqn.~\ref{eqn:dirichletmaster} implies that $\frac{d}{d\bm{\alpha}} \sum_i z_i = 0$, as it must because of the simplex constraint.
Since we have already developed an approximation for $\frac{\partial F_{\rm{Beta}}}{\partial \theta}$, Eqn.~\ref{eqn:dirichletmaster}
provides a complete recipe for pathwise Dirichlet gradients. Note that although we have used a stick-breaking
construction to derive Eqn.~\ref{eqn:dirichletmaster}, this in no way dictates the sampling scheme we use when
generating $\bm{z} \sim \rm{Dir}(\bm{\alpha})$. In the supplementary materials we verify that Eqn.~\ref{eqn:dirichletmaster}
satisfies the transport equation.

\subsection{Implementation}

It is worth emphasizing that pathwise gradient estimators of the form in Eqn.~\ref{eqn:multiest} have the
advantage of being `plug-and-play.' We simply plug an approximate or exact velocity field into our
favorite automatic differentiation engine\footnote{Our approximations for pathwise gradients for 
the Gamma, Beta, and Dirichlet distributions are available in the 0.4 release of PyTorch \cite{paszke2017automatic}.} so that samples $\bm{z}$ and $f_\bth(\bm{z})$ are differentiable
w.r.t.~$\bth$. There is no need to construct a surrogate objective
function to form the gradient estimator.

\section{Related Work}
\label{sec:relatedwork}

A number of lines of research bears upon our work. There is a large body of work on constructing
gradient estimators with reduced variance, much of which can be understood in terms of control variates \cite{ross}:
for example, \cite{mnih2014neural} construct neural baselines for score-function gradients; \cite{schulman2015gradient}
discuss gradient estimators for stochastic computation graphs and their Rao-Blackwellization;
 and \cite{tucker2017rebar, grathwohl2017backpropagation}
construct adaptive control variates for discrete random variables. Another example of this line of work is reference \cite{miller2017reducing},
where the authors construct control variates that are applicable when $q_\bth(\bz)$ is a \emph{diagonal} Normal distribution. While our OMT
gradient for the multivariate Normal distribution, Eqn.~\ref{eqn:mvnopt}, can also be understood in the language of control 
variates,\footnote{See Sec.~\ref{sec:discussion} and the supplementary materials for a brief discussion.} \cite{miller2017reducing}
relies on Taylor expansions of the test function $f_\bth(\bz)$.\footnote{In addition, note that in their approach variance reduction for gradients w.r.t.~the scale
parameter $\bm{\sigma}$ necessitates a multi-sample estimator (at least for high-dimensional models where computing the diagonal of the Hessian is prohibitively expensive).}

In \cite{graves2016stochastic}, the author derives formula Eqn.~\ref{eqn:master} and uses it to construct gradient estimators for mixture distributions. 
Unfortunately, the resulting gradient estimator is expensive, relying on a recursive computation that scales with the dimension of the sample space. 

\begin{figure}[t]
\begin{center}
\centerline{\includegraphics[width=.8\columnwidth]{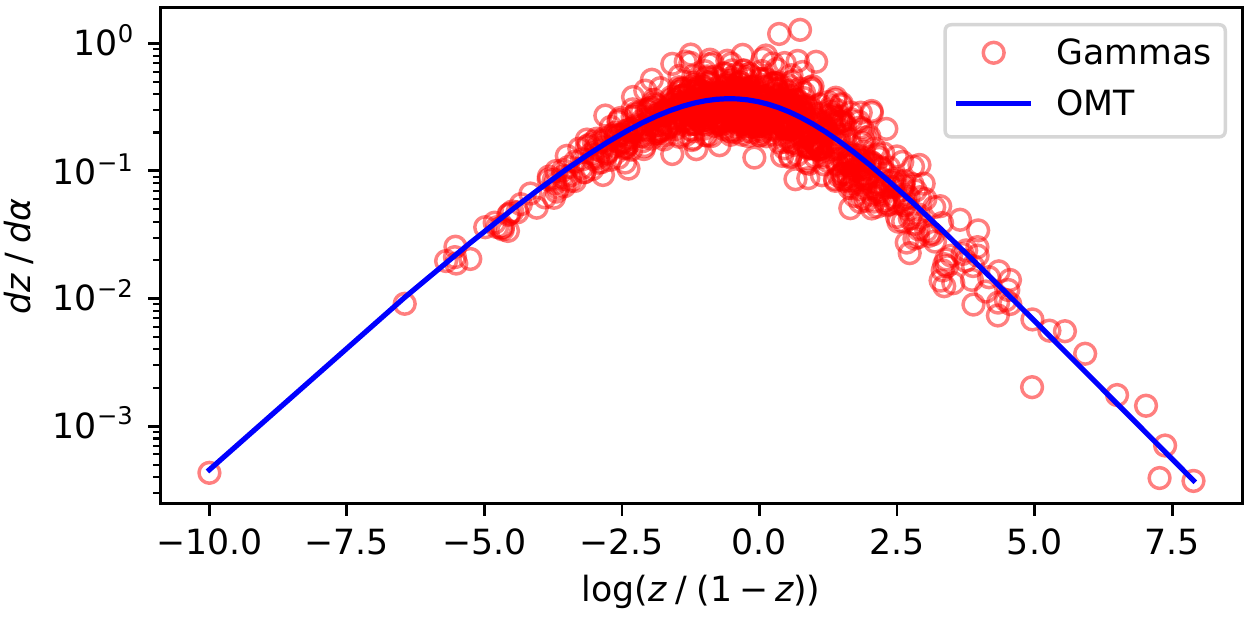}}
\caption{Derivatives $\frac{d z}{d \alpha}$ for samples $z \sim {\rm Beta}(1,1)$. We compare the OMT gradient to the gradient
that is obtained when samples $z \sim$ Beta($\alpha, \beta$) are represented as the ratio of two Gamma variates (each with its own pathwise derivative).
The OMT derivative has a deterministic value for each sample $z$, whereas the Gamma representation induces a higher variance stochastic derivative due to the presence of an auxiliary random variable.}
\label{fig:betagamma}
\end{center}
\end{figure}

Another line of work constructs partially reparameterized gradient estimators for cases where the reparameterization
trick is difficult to apply.
The \gls{G-Rep} \cite{ruiz2016generalized} uses standardization via sufficient
statistics to obtain 
a transformation $\mathcal{T}(\beps; \bth)$ that minimizes the dependence of $q(\beps)$ on $\bth$. This
results in a partially reparameterized gradient estimator that also includes a score function-like term.\footnote{That is a term
in the gradient estimator that is proportional to the test function $f_\bth(\bz)$.} 
In \acrshort{RSVI} \cite{naesseth2017reparameterization} the authors consider gradient estimators in the case that $q_\bth(\bz)$
can be sampled from efficiently via rejection sampling. This results in a gradient estimator with the same 
generic structure as \acrshort{G-Rep}, although in the case of \acrshort{RSVI} the score function-like term can often be dropped 
in practice at the cost of small bias (with the benefit of reduced variance). Besides the fact that this gradient estimator
is not fully pathwise, one key difference with our approach is that for many distributions of interest (e.g.~the Beta
and Dirichlet distributions), rejection sampling introduces auxiliary random variables, which results in additional stochasticity and thus
higher variance (cf.~Figure \ref{fig:betagamma}). In contrast our pathwise gradients for the Beta and Dirichlet distributions are \emph{deterministic} for a given $\bz$ and $\bth$. Finally, \cite{knowles2015stochastic} uses (somewhat imprecise) approximations to the inverse CDF to derive gradient estimators for Gamma random variables.

As the final version of this manuscript was being prepared, we became aware of \cite{figurnov2018implicit}, which has some overlap with this
work. In particular, \cite{figurnov2018implicit} derives Eqn.~\ref{eqn:master} and an interesting
 generalization to the multivariate case. This allows the authors
to construct pathwise derivatives for the Gamma, Beta, and Dirichlet distributions. For the latter two distributions, however, the derivatives include
additional stochasticity that our pathwise derivatives avoid. Also, the authors do not draw the connection to the transport equation and optimal
transport or consider
 the multivariate Normal distribution in any detail.

\vspace{-3pt}

\section{Experiments}
\label{sec:exp}

All experiments in this section use single-sample gradient estimators.

\subsection{Synthetic Experiments}

In this section we validate our pathwise gradients for the Beta, Dirichlet, and multivariate Normal distributions.
Where appropriate we compare to the RT gradient, the score function gradient, or \acrshort{RSVI}.

\subsubsection{Beta Distribution}
\label{sec:betasynth}
\begin{figure}[t]
\begin{center}
\centerline{\includegraphics[width=.9\columnwidth]{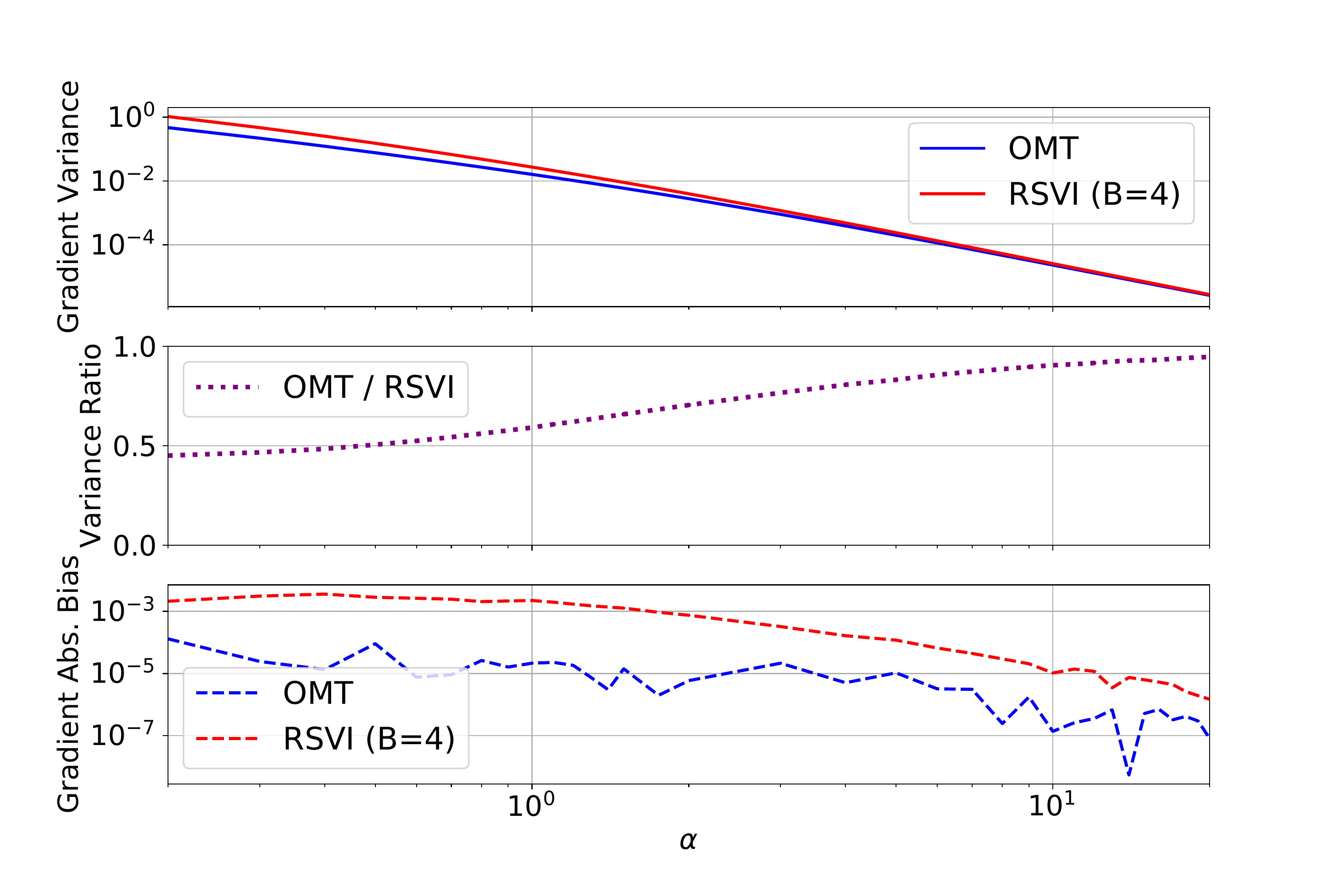}}
\caption{We compare the OMT gradient to the \acrshort{RSVI} gradient with $B=4$
for the test function $f(z) = z^3$ and $q_\bth(z) = {\rm Beta}(z| \alpha, \alpha)$. In the bottom panel we depict finite-sample bias
for 25 million samples (this also includes effects from finite numerical precision).}
\label{fig:beta}
\end{center}
\end{figure}

In Fig.~\ref{fig:beta} we compare the performance of our OMT gradient for Beta random variables to the \acrshort{RSVI}
gradient estimator. We use a test function $f(z)=z^3$ for which we can compute the gradient exactly. We see that
the OMT gradient performs favorably over the entire range of parameter $\alpha$ that defines 
the distribution ${\rm Beta}(\alpha, \alpha)$ used to compute $\mathcal{L}$. For smaller $\alpha$, where $\mathcal{L}$ exhibits
larger curvature, the variance of the estimator is noticeably reduced. Notice that one reason for the reduced variance of the OMT
estimator as compared to the \acrshort{RSVI} estimator is the presence of an auxiliary random variable in the latter case (cf.~Figure \ref{fig:betagamma}).

\vspace{-3pt}

\subsubsection{Dirichlet Distribution}
\label{sec:diri}
\begin{figure}[t]
\begin{center}
\centerline{\includegraphics[width=\columnwidth]{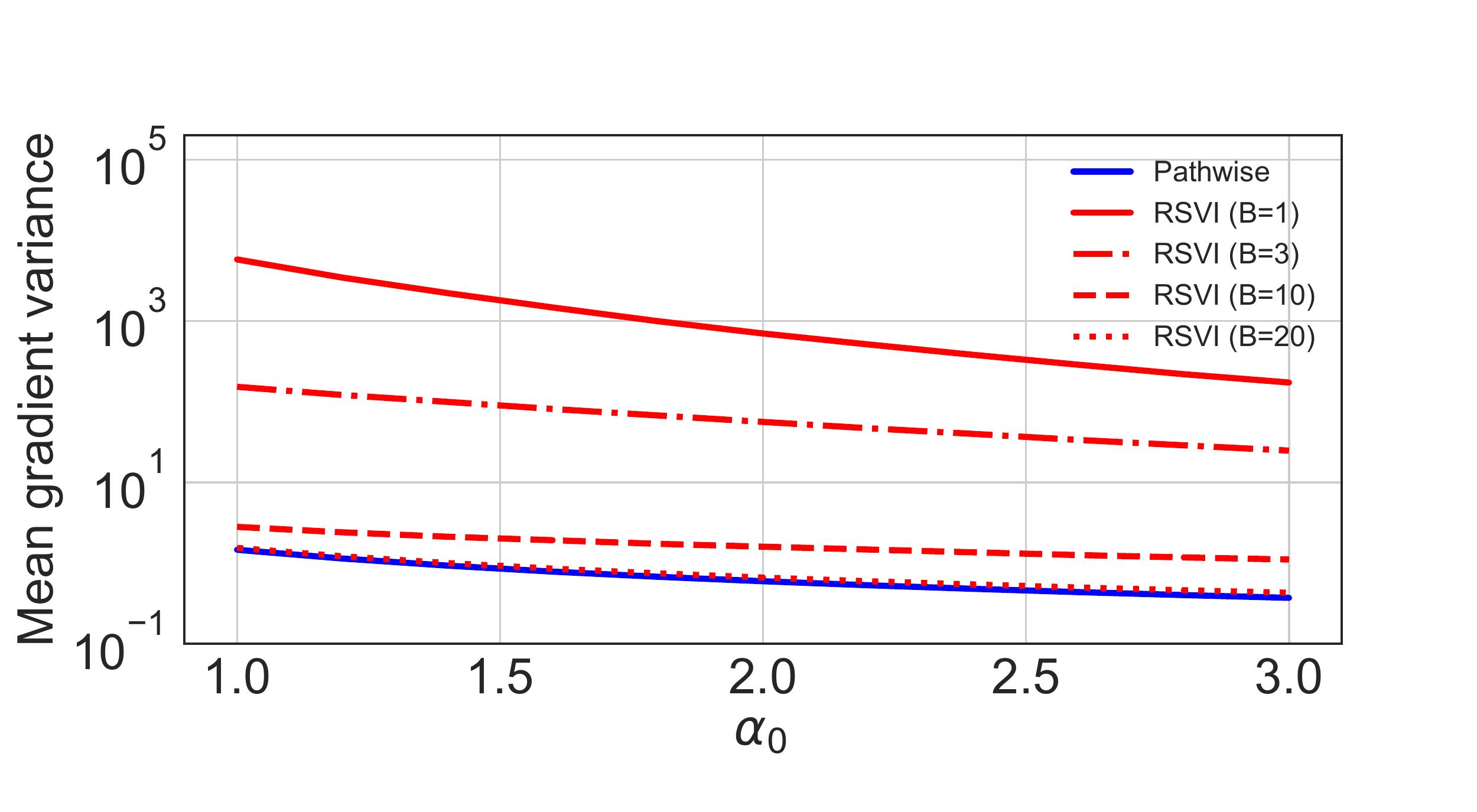}}
\caption{Gradient variance for the ELBO of a conjugate Multinomial-Dirichlet model. 
We compare the pathwise gradient to \acrshort{RSVI} for different boosts $B$. See Sec.~\ref{sec:diri} for details.}
\label{fig:dirichlet}
\end{center}
\end{figure}

In Fig.~\ref{fig:dirichlet} we compare the variance of our pathwise gradient for the
Dirichlet distribution to the \acrshort{RSVI}
gradient estimator. We compute stochastic gradients of the ELBO for a Multinomial-Dirichlet model initialized at the exact
posterior (where the exact gradient is zero). The Dirichlet distribution has 1995 components, and the single data point is a bag of words from 
a natural language document. We see that
the pathwise gradient performs favorably over the entire range of the model hyperparameter $\alpha_0$ considered.
Note that as we crank up the shape augmentation setting $B$, the \acrshort{RSVI} variance approaches that of the pathwise 
gradient.\footnote{As discussed in Sec.~\ref{sec:relatedwork}, the variance of the \acrshort{RSVI} gradient estimator can also be reduced
by dropping the score function-like term (at the cost of some bias).}

\vspace{-3pt}

\subsubsection{Multivariate Normal}

\begin{figure}[t]
\begin{center}
  \centerline{\includegraphics[width=.5 \textwidth]{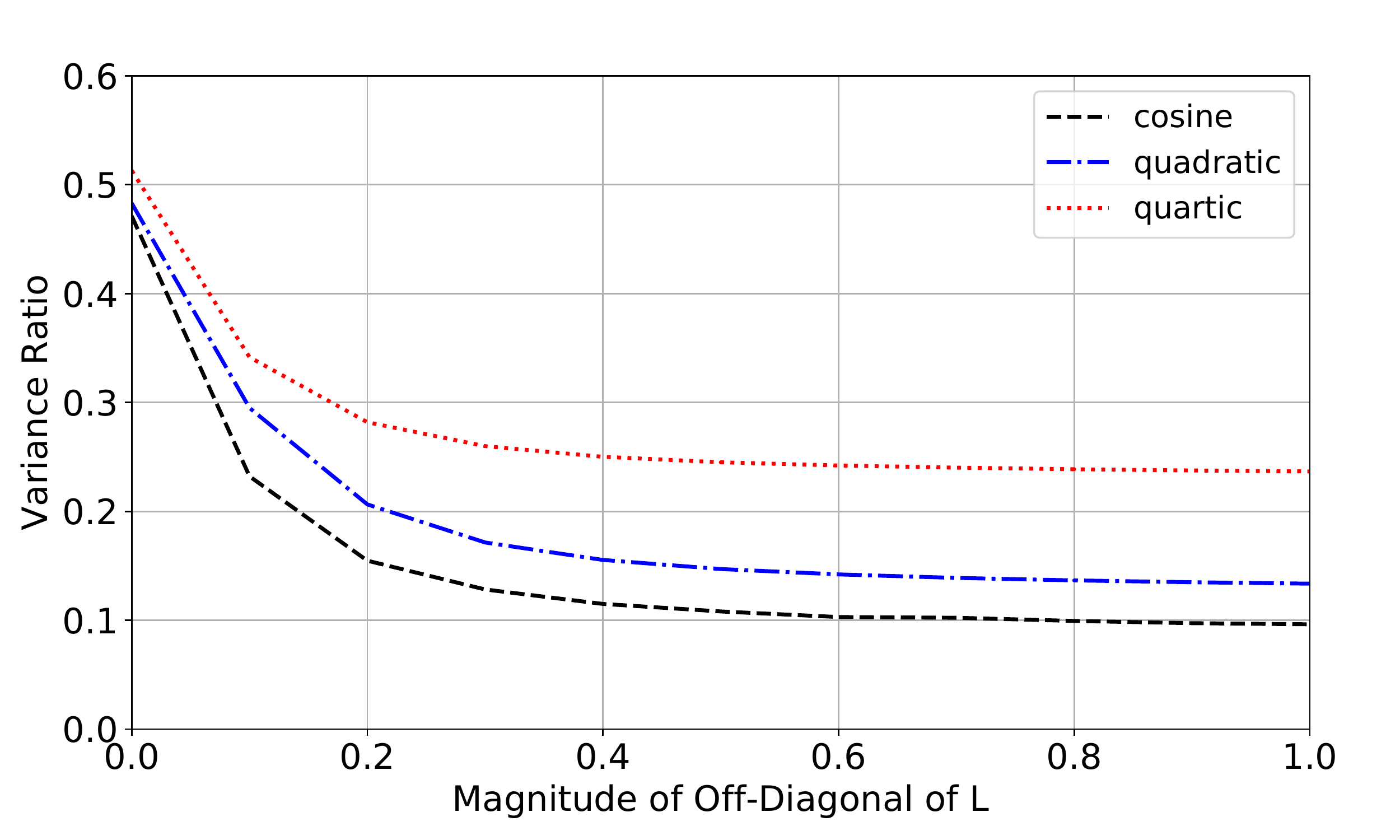}}
        \caption{We compare the OMT gradient estimator for the multivariate Normal distribution to the RT estimator for three test functions.
        The horizontal axis controls the magnitude of the off-diagonal elements of the Cholesky factor $\bm{L}$.
The vertical axis depicts the ratio of the mean variance of the OMT estimator to that of the RT estimator for the off-diagonal elements of $\bm{L}$.}
\label{fig:omtmvnreach}
\end{center}
\end{figure}

In Fig.~\ref{fig:omtmvnreach} we use synthetic test functions to illustrate the amount of variance reduction that can be achieved
with the OMT gradient estimator for the multivariate Normal distribution. 
The dimension is $D=50$; the results are qualitatively similar for different dimensions.

\vspace{-3pt}

\subsection{Real World Datasets}

In this section we investigate the performance of our gradient estimators for the Gamma, Beta, and multivariate Normal
distributions in two variational inference tasks on
real world datasets. Note that we include an additional experiment for the multivariate Normal distribution in the supplementary
materials, see Sec.~9.11.
All the experiments in this section were implemented in the Pyro\footnote{\url{http://pyro.ai}} probabilistic programming language.

\subsubsection{Sparse Gamma \acrshort{DEF}}

\begin{figure}[t]
\begin{center}
\centerline{\includegraphics[width=\columnwidth]{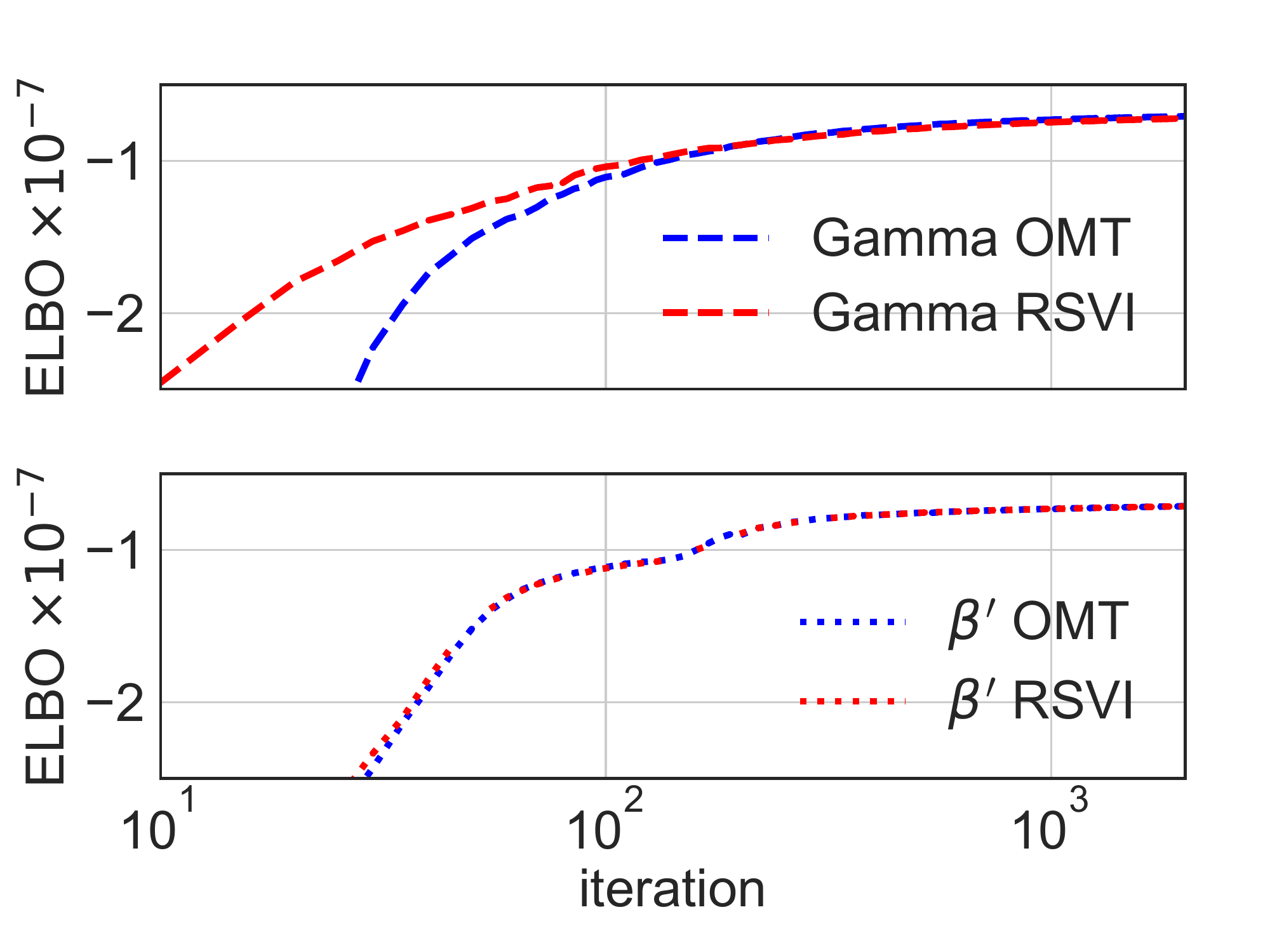}}
\caption{ELBO during training for two variants of the Sparse Gamma \acrshort{DEF}, one with and one without
Beta random variables. We compare the OMT gradient to \acrshort{RSVI}. At each iteration we depict a multi-sample estimate of the ELBO
with $N=100$ samples.}
\label{fig:def}
\end{center}
\end{figure}

The Sparse Gamma \acrshort{DEF} \cite{ranganath2015deep} is a probabilistic model with multiple layers of local latent
random variables $z_{nk}^{(\ell)}$ and global random weights $w_{k k^\prime}^{(\ell)}$ that mimics the architecture of a deep neural network. Here each $n$ corresponds to an observed data point $x_n$, $\ell$ indexes the layer, and $k$ and $k^\prime$ run over the
latent components. 
We consider Poisson-distributed observations $x_{nd}$ for each dimension $d$. Concretely, the model is specified as\footnote{
Note that this experiment closely follows the setup in \cite{ruiz2016generalized} and \cite{naesseth2017reparameterization}.}
\vspace{-2pt}
\begin{equation}
\nonumber
\begin{split}
     &z_{nk}^{(\ell)} \sim \textrm{Gamma}\left(\alpha_z, \frac{\alpha_z}{\sum_{k^\prime} z_{nk^\prime}^{(\ell+1)} w_{k^\prime k}^{(\ell)}}\right)
     \qquad \ell=1,2,...,L-1 \\
    &x_{nd} \sim \textrm{Poisson}\left(\sum_{k^\prime} z_{nk^\prime}^{(1)} w_{k^\prime d}^{(0)}\right) \qquad 
    z_{nk}^{L} \sim \textrm{Gamma}\left(\alpha_z, \alpha_z \right)
\end{split}
\end{equation}
We  set $\alpha_z = 0.1$ and use $L=3$ layers with $100$, $40$, and $15$ latent factors per 
data point (for $\ell = 1,2,3$, respectively).
We consider two model variants that differ in the prior placed on the weights.
In the first variant we place Gamma priors over the weights with $\alpha = 0.3$ and $\beta = 0.1$.
 In the second variant we place $\beta^\prime$ priors over the weights with the same means
and variances as in the first variant.\footnote{If $z \sim {\rm Beta}(\alpha, \beta)$ then $\frac{z}{1-z} \sim {\rm \beta^\prime}(\alpha, \beta)$. Thus like the Gamma distribution the Beta prime distribution has support on the positive real line.} 
The dataset we consider is the Olivetti faces dataset,\footnote{\url{http://www.cl.cam.ac.uk/research/dtg/attarchive/facedatabase.html}}
which consists of $64\times 64$ grayscale images of human faces. 
In Fig.~\ref{fig:def} we depict how the training set ELBO increases during the course of optimization. We find that on
this task the performance of the OMT gradient estimator is nearly identical to \acrshort{RSVI}.\footnote{Note
that we do not compare to any alternative estimators such as \acrshort{G-Rep}, since
\cite{naesseth2017reparameterization} shows that \acrshort{RSVI} has superior performance on this task.} 
Figure \ref{fig:def} suggests that gradient variance is not the limiting factor for this particular task and dataset.

\vspace{-3pt}

\subsubsection{Gaussian Process Regression}
\label{sec:gp}

\begin{figure}[ht]
\begin{center}
\centerline{\includegraphics[width=\columnwidth]{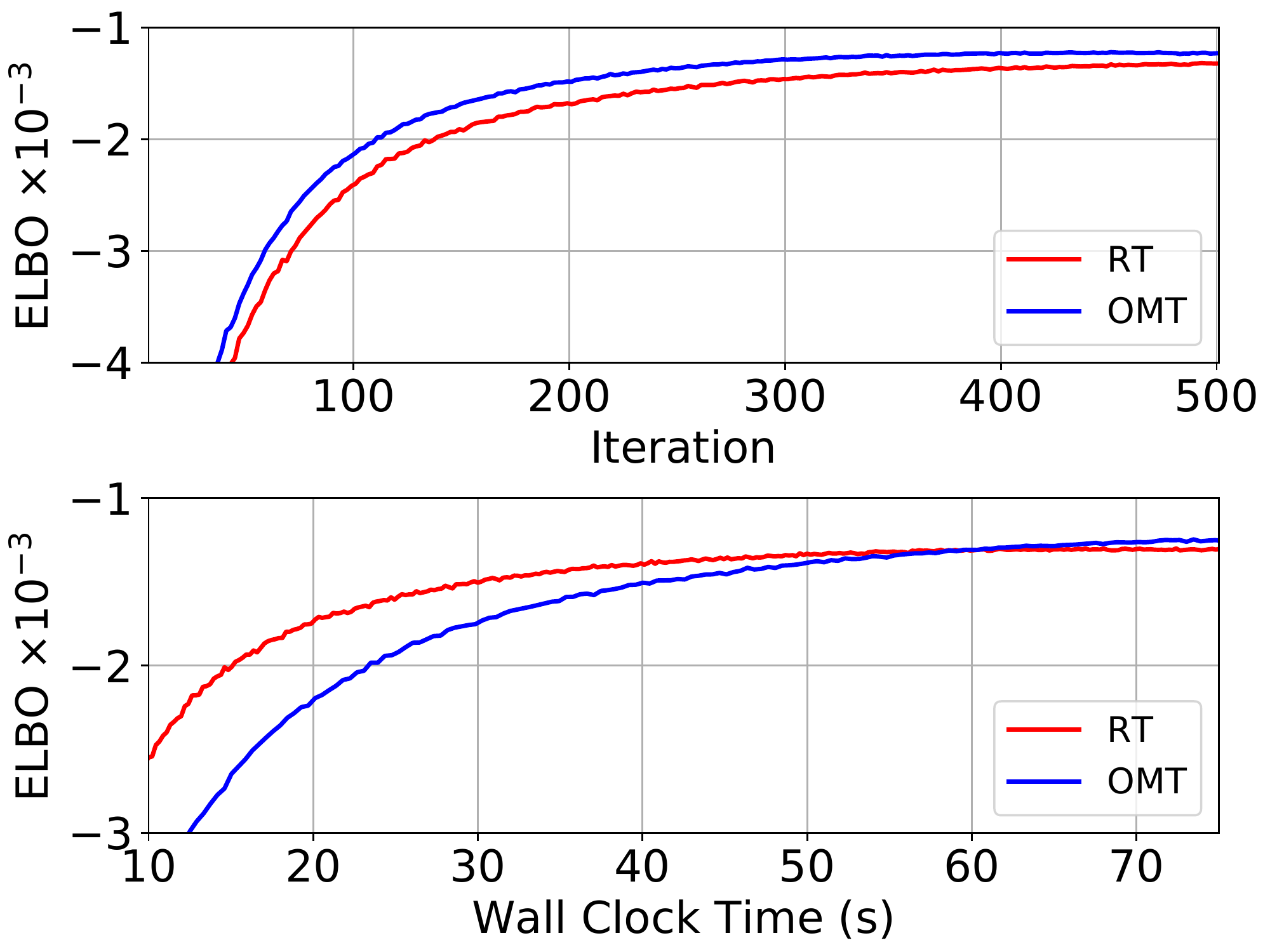}}
\caption{ELBO during training for the Gaussian Process regression task in Sec.~\ref{sec:gp}.
 At each iteration we depict a multi-sample estimate of the ELBO
with $N=100$ samples. We compare the OMT gradient estimator to the RT estimator.}
\label{fig:gp}
\end{center}
\end{figure}

In this section we investigate the performance of our OMT gradient for the multivariate Normal
distribution, Eqn.~\ref{eqn:mvnopt}, in the context of a Gaussian Process regression task. 
We model the Mauna Loa ${\rm CO}_2$ data from \cite{keeling2004atmospheric} considered
in \cite{rasmussen2004gaussian}. We use a structured kernel that accommodates a long term linear trend as well as a
periodic component. We fit the GP using a single-sample Monte Carlo ELBO gradient estimator and all $D=468$ data points.
The variational family is a multivariate Normal distribution with a Cholesky parameterization for the covariance matrix.
Progress on the ELBO during the course of training is depicted in Fig.~\ref{fig:gp}. We can see that the OMT gradient
estimator has superior sample efficiency due to its lower variance. By iteration 270 the OMT gradient
estimator has attained the same ELBO that the RT estimator attains at iteration 500. Since each iteration of the OMT
estimator is $\sim\!1.9$x slower than the corresponding RT iteration, the superior sample efficiency of the OMT estimator
is largely canceled when judged by wall clock time. Nevertheless, the lower variance of the OMT estimator results
in a higher ELBO than that obtained by the RT estimator.

\vspace{-3pt}

\section{Discussion and Future Work}
\label{sec:discussion}

We have seen that optimal transport offers a fruitful perspective on pathwise gradients. On the one hand
it has helped us formulate pathwise gradients in situations where this was assumed to be impractical.
On the other hand it has focused our attention on a particular notion of optimality, which
led us to develop a new gradient estimator for the multivariate Normal distribution. 
A better understanding of this notion of optimality and, more broadly, a better understanding of when pathwise gradients
are preferable over score function gradients (or vice versa) would be useful in guiding the practical application of these methods.

Since each solution of the transport equation Eqn.~\ref{eqn:transport} yields an unbiased gradient estimator,
the difference between any two such estimators can be thought of as a control variate. In the case
of the multivariate Normal distribution, where computing the OMT gradient has a cost $\mathcal{O}(D^{3})$, an
attractive alternative to using $\bm{v}^{\rm OMT}$ is to adaptively choose $\bm{v}$ during the course of optimization
in direct analogy to adaptive control variate techniques. 
In future work we will explore this approach in detail, which promises lower variance
than the OMT estimator at reduced computational cost.

The geometric picture from optimal transport---and thus the potential for non-trivial derivative applications---is especially rich for multivariate distributions. 
Here we have explored the multivariate Normal and Dirichlet distributions in some detail, but this just scratches the surface
of multivariate distributions. It would be of general interest to develop pathwise gradients for a broader class of multivariate distributions, including for example mixture distributions. Rich distributions with low variance gradient estimators are of special interest in the context of SGVI, where the need to approximate complex posteriors demands rich families of distributions that lend themselves to stochastic optimization. In future work we intend to explore this connection further.

\section*{Acknowledgements}

We thank Peter Dayan and Zoubin Ghahramani for feedback on a draft manuscript
and other colleagues at Uber AI Labs---especially Noah Goodman and Theofanis Karaletsos---
for stimulating conversations during the course of this work. We also thank Christian Naesseth
for clarifying details of the experimental setup for the deep exponential family experiment in \cite{naesseth2017reparameterization}.


\bibliography{gamma_beta}

\begin{thebibliography}{41}
\providecommand{\natexlab}[1]{#1}
\providecommand{\url}[1]{\texttt{#1}}
\expandafter\ifx\csname urlstyle\endcsname\relax
  \providecommand{\doi}[1]{doi: #1}\else
  \providecommand{\doi}{doi: \begingroup \urlstyle{rm}\Url}\fi

\bibitem[Ambrosio et~al.(2008)Ambrosio, Gigli, and
  Savar{\'e}]{ambrosio2008gradient}
Ambrosio, Luigi, Gigli, Nicola, and Savar{\'e}, Giuseppe.
\newblock \emph{Gradient flows: in metric spaces and in the space of
  probability measures}.
\newblock Springer Science \& Business Media, 2008.

\bibitem[Butler(2007)]{butler2007saddlepoint}
Butler, Ronald~W.
\newblock \emph{Saddlepoint approximations with applications}, volume~22.
\newblock Cambridge University Press, 2007.

\bibitem[Casella \& Robert(1996)Casella and Robert]{casella1996rao}
Casella, George and Robert, Christian~P.
\newblock Rao-blackwellisation of sampling schemes.
\newblock \emph{Biometrika}, 83\penalty0 (1):\penalty0 81--94, 1996.

\bibitem[Duchi et~al.(2011)Duchi, Hazan, and Singer]{duchi2011adaptive}
Duchi, John, Hazan, Elad, and Singer, Yoram.
\newblock Adaptive subgradient methods for online learning and stochastic
  optimization.
\newblock \emph{Journal of Machine Learning Research}, 12\penalty0
  (Jul):\penalty0 2121--2159, 2011.

\bibitem[Efron \& Morris(1975)Efron and Morris]{efron1975data}
Efron, Bradley and Morris, Carl.
\newblock Data analysis using stein's estimator and its generalizations.
\newblock \emph{Journal of the American Statistical Association}, 70\penalty0
  (350):\penalty0 311--319, 1975.

\bibitem[Figurnov et~al.(2018)Figurnov, Mohamed, and
  Mnih]{figurnov2018implicit}
Figurnov, Michael, Mohamed, Shakir, and Mnih, Andriy.
\newblock Implicit reparameterization gradients.
\newblock \emph{arXiv preprint arXiv:1805.08498}, 2018.

\bibitem[Glasserman(2013)]{glasserman2013monte}
Glasserman, Paul.
\newblock \emph{Monte Carlo methods in financial engineering}, volume~53.
\newblock Springer Science \& Business Media, 2013.

\bibitem[Glynn(1990)]{glynn1990likelihood}
Glynn, Peter~W.
\newblock Likelihood ratio gradient estimation for stochastic systems.
\newblock \emph{Communications of the ACM}, 33\penalty0 (10):\penalty0 75--84,
  1990.

\bibitem[Grathwohl et~al.(2017)Grathwohl, Choi, Wu, Roeder, and
  Duvenaud]{grathwohl2017backpropagation}
Grathwohl, Will, Choi, Dami, Wu, Yuhuai, Roeder, Geoff, and Duvenaud, David.
\newblock Backpropagation through the void: Optimizing control variates for
  black-box gradient estimation.
\newblock \emph{arXiv preprint arXiv:1711.00123}, 2017.

\bibitem[Graves(2016)]{graves2016stochastic}
Graves, Alex.
\newblock Stochastic backpropagation through mixture density distributions.
\newblock \emph{arXiv preprint arXiv:1607.05690}, 2016.

\bibitem[Hoffman et~al.(2013)Hoffman, Blei, Wang, and
  Paisley]{hoffman2013stochastic}
Hoffman, Matthew~D, Blei, David~M, Wang, Chong, and Paisley, John.
\newblock Stochastic variational inference.
\newblock \emph{The Journal of Machine Learning Research}, 14\penalty0
  (1):\penalty0 1303--1347, 2013.

\bibitem[Jordan et~al.(1999)Jordan, Ghahramani, Jaakkola, and
  Saul]{jordan1999introduction}
Jordan, Michael~I, Ghahramani, Zoubin, Jaakkola, Tommi~S, and Saul, Lawrence~K.
\newblock An introduction to variational methods for graphical models.
\newblock \emph{Machine learning}, 37\penalty0 (2):\penalty0 183--233, 1999.

\bibitem[Keeling \& Whorf(2004)Keeling and Whorf]{keeling2004atmospheric}
Keeling, Charles~David and Whorf, Timothy~P.
\newblock Atmospheric co2 concentrations derived from flask air samples at
  sites in the sio network.
\newblock \emph{Trends: a compendium of data on Global Change}, 2004.

\bibitem[Kingma \& Ba(2014)Kingma and Ba]{kingma2014adam}
Kingma, Diederik~P and Ba, Jimmy.
\newblock Adam: A method for stochastic optimization.
\newblock \emph{arXiv preprint arXiv:1412.6980}, 2014.

\bibitem[Kingma \& Welling(2013)Kingma and Welling]{kingma2013auto}
Kingma, Diederik~P and Welling, Max.
\newblock Auto-encoding variational bayes.
\newblock \emph{arXiv preprint arXiv:1312.6114}, 2013.

\bibitem[Knowles(2015)]{knowles2015stochastic}
Knowles, David~A.
\newblock Stochastic gradient variational bayes for gamma approximating
  distributions.
\newblock \emph{arXiv preprint arXiv:1509.01631}, 2015.

\bibitem[Kucukelbir et~al.(2016)Kucukelbir, Tran, Ranganath, Gelman, and
  Blei]{kucukelbir2016automatic}
Kucukelbir, Alp, Tran, Dustin, Ranganath, Rajesh, Gelman, Andrew, and Blei,
  David~M.
\newblock Automatic differentiation variational inference.
\newblock \emph{arXiv preprint arXiv:1603.00788}, 2016.

\bibitem[Lugannani \& Rice(1980)Lugannani and Rice]{lugannani1980saddle}
Lugannani, Robert and Rice, Stephen.
\newblock Saddle point approximation for the distribution of the sum of
  independent random variables.
\newblock \emph{Advances in applied probability}, 12\penalty0 (2):\penalty0
  475--490, 1980.

\bibitem[Miller et~al.(2017)Miller, Foti, D'Amour, and
  Adams]{miller2017reducing}
Miller, Andrew, Foti, Nick, D'Amour, Alexander, and Adams, Ryan~P.
\newblock Reducing reparameterization gradient variance.
\newblock In \emph{Advances in Neural Information Processing Systems}, pp.\
  3711--3721, 2017.

\bibitem[Mnih \& Gregor(2014)Mnih and Gregor]{mnih2014neural}
Mnih, Andriy and Gregor, Karol.
\newblock Neural variational inference and learning in belief networks.
\newblock \emph{arXiv preprint arXiv:1402.0030}, 2014.

\bibitem[Naesseth et~al.(2017)Naesseth, Ruiz, Linderman, and
  Blei]{naesseth2017reparameterization}
Naesseth, Christian, Ruiz, Francisco, Linderman, Scott, and Blei, David.
\newblock Reparameterization gradients through acceptance-rejection sampling
  algorithms.
\newblock In \emph{Artificial Intelligence and Statistics}, pp.\  489--498,
  2017.

\bibitem[Paisley et~al.(2012)Paisley, Blei, and Jordan]{paisley2012variational}
Paisley, John, Blei, David, and Jordan, Michael.
\newblock Variational bayesian inference with stochastic search.
\newblock \emph{arXiv preprint arXiv:1206.6430}, 2012.

\bibitem[Paszke et~al.(2017)Paszke, Gross, Chintala, Chanan, Yang, DeVito, Lin,
  Desmaison, Antiga, and Lerer]{paszke2017automatic}
Paszke, Adam, Gross, Sam, Chintala, Soumith, Chanan, Gregory, Yang, Edward,
  DeVito, Zachary, Lin, Zeming, Desmaison, Alban, Antiga, Luca, and Lerer,
  Adam.
\newblock Automatic differentiation in pytorch.
\newblock 2017.

\bibitem[Price(1958)]{price1958useful}
Price, Robert.
\newblock A useful theorem for nonlinear devices having gaussian inputs.
\newblock \emph{IRE Transactions on Information Theory}, 4\penalty0
  (2):\penalty0 69--72, 1958.

\bibitem[Ranganath et~al.(2014)Ranganath, Gerrish, and
  Blei]{ranganath2014black}
Ranganath, Rajesh, Gerrish, Sean, and Blei, David.
\newblock Black box variational inference.
\newblock In \emph{Artificial Intelligence and Statistics}, pp.\  814--822,
  2014.

\bibitem[Ranganath et~al.(2015)Ranganath, Tang, Charlin, and
  Blei]{ranganath2015deep}
Ranganath, Rajesh, Tang, Linpeng, Charlin, Laurent, and Blei, David.
\newblock Deep exponential families.
\newblock In \emph{Artificial Intelligence and Statistics}, pp.\  762--771,
  2015.

\bibitem[Rasmussen(2004)]{rasmussen2004gaussian}
Rasmussen, Carl~Edward.
\newblock Gaussian processes in machine learning.
\newblock In \emph{Advanced lectures on machine learning}, pp.\  63--71.
  Springer, 2004.

\bibitem[Rezende et~al.(2014)Rezende, Mohamed, and
  Wierstra]{rezende2014stochastic}
Rezende, Danilo~Jimenez, Mohamed, Shakir, and Wierstra, Daan.
\newblock Stochastic backpropagation and approximate inference in deep
  generative models.
\newblock \emph{arXiv preprint arXiv:1401.4082}, 2014.

\bibitem[Robbins \& Monro(1951)Robbins and Monro]{robbins1951stochastic}
Robbins, Herbert and Monro, Sutton.
\newblock A stochastic approximation method.
\newblock \emph{The annals of mathematical statistics}, pp.\  400--407, 1951.

\bibitem[Ross(2006)]{ross}
Ross, Sheldon~M.
\newblock \emph{Simulation}.
\newblock Academic Press, San Diego, 2006.

\bibitem[Ruiz et~al.(2016)Ruiz, AUEB, and Blei]{ruiz2016generalized}
Ruiz, Francisco~R, AUEB, Michalis Titsias~RC, and Blei, David.
\newblock The generalized reparameterization gradient.
\newblock In \emph{Advances in Neural Information Processing Systems}, pp.\
  460--468, 2016.

\bibitem[Salimans et~al.(2013)Salimans, Knowles, et~al.]{salimans2013fixed}
Salimans, Tim, Knowles, David~A, et~al.
\newblock Fixed-form variational posterior approximation through stochastic
  linear regression.
\newblock \emph{Bayesian Analysis}, 8\penalty0 (4):\penalty0 837--882, 2013.

\bibitem[Schulman et~al.(2015)Schulman, Heess, Weber, and
  Abbeel]{schulman2015gradient}
Schulman, John, Heess, Nicolas, Weber, Theophane, and Abbeel, Pieter.
\newblock Gradient estimation using stochastic computation graphs.
\newblock In \emph{Advances in Neural Information Processing Systems}, pp.\
  3528--3536, 2015.

\bibitem[{Stan Manual}(2017)]{stanmanual}
{Stan Manual}.
\newblock Stan modeling language users guide and reference manual, version
  2.17.0.
\newblock
  \emph{\url{http://mc-stan.org/users/documentation/case-studies/pool-binary-trials.html}},
  2017.

\bibitem[Tieleman \& Hinton(2012)Tieleman and Hinton]{Tieleman2012}
Tieleman, T. and Hinton, G.
\newblock {Lecture 6.5---RmsProp: Divide the gradient by a running average of
  its recent magnitude}.
\newblock COURSERA: Neural Networks for Machine Learning, 2012.

\bibitem[Titsias \& L{\'a}zaro-Gredilla(2014)Titsias and
  L{\'a}zaro-Gredilla]{titsias2014doubly}
Titsias, Michalis and L{\'a}zaro-Gredilla, Miguel.
\newblock Doubly stochastic variational bayes for non-conjugate inference.
\newblock In \emph{International Conference on Machine Learning}, pp.\
  1971--1979, 2014.

\bibitem[Tucker et~al.(2017)Tucker, Mnih, Maddison, Lawson, and
  Sohl-Dickstein]{tucker2017rebar}
Tucker, George, Mnih, Andriy, Maddison, Chris~J, Lawson, John, and
  Sohl-Dickstein, Jascha.
\newblock Rebar: Low-variance, unbiased gradient estimates for discrete latent
  variable models.
\newblock In \emph{Advances in Neural Information Processing Systems}, pp.\
  2624--2633, 2017.

\bibitem[Villani(2003)]{villani2003topics}
Villani, C{\'e}dric.
\newblock \emph{Topics in optimal transportation}.
\newblock Number~58. American Mathematical Soc., 2003.

\bibitem[Wilks(1962)]{wilks}
Wilks, S.S.
\newblock \emph{Mathematical Statistics}.
\newblock John Wiley and Sons Inc., 1962.

\bibitem[Williams(1992)]{williams1992simple}
Williams, Ronald~J.
\newblock Simple statistical gradient-following algorithms for connectionist
  reinforcement learning.
\newblock \emph{Machine learning}, 8\penalty0 (3-4):\penalty0 229--256, 1992.

\bibitem[Wingate \& Weber(2013)Wingate and Weber]{wingate2013automated}
Wingate, David and Weber, Theophane.
\newblock Automated variational inference in probabilistic programming.
\newblock \emph{arXiv preprint arXiv:1301.1299}, 2013.

\end{thebibliography}
\bibliographystyle{icml2018}

\section{Supplementary Materials}

\subsection{The Univariate Case}

For completeness we show explicitly that the formula
\begin{equation}
\label{eqn:mf}
\frac{dz}{d\theta}  = -\frac{\frac{\partial F_\bth}{\partial \theta }(z)}{q_\bth(z)}
\end{equation}
yields the correct gradient. Without loss of generality we assume that $f(z)$
has no explicit dependence on $\theta$.
Substituting Eqn.~\ref{eqn:mf} for $\frac{dz}{d\theta}$ we have 
\begin{equation}
\label{eqn:1dderiv}
\begin{split}
\E_{q_\bth(z)}\! \left[ \frac{\partial f}{\partial z} \frac{\partial z}{\partial \theta}  \right] &= 
-\! \int_{-\infty}^{\infty} \! \frac{q_\bth(z)}{q_\bth(z)} \frac{\partial f}{\partial z}  \int_{-\infty}^{z}\! 
\frac{\partial q_\bth(z^\prime)}{\partial \theta}  dz^\prime \!dz \\
&= -\! \int_{-\infty}^{\infty} \! 
\frac{\partial q_\bth(z^\prime)}{\partial \theta} \! \int_{z^\prime}^{\infty}\! \frac{\partial f}{\partial z}  dz dz^\prime \\
&= -\! \int_{-\infty}^{\infty} \! 
\frac{\partial q_\bth(z^\prime)}{\partial \theta} \! \left( -f(z^\prime) \right) dz^\prime \\
&= \frac{d}{d\theta} E_{q_\bth(z)}\! \left[ f(z)  \right]
\end{split}
\end{equation}
In the second line we changed the order of integration and in the third we appealed
to the fundamental theorem of calculus, assuming that $f(z)$ is sufficiently
regular that we can drop the boundary term at infinity.

Note that Eqn.~\ref{eqn:mf} is the unique solution $v= \frac{dz}{d\theta}$ to the one-dimensional version of the transport
equation that satisfies the boundary condition $\lim_{z\to\infty} q_\bth v=0$:
\begin{equation}
\label{eqn:1dtransport}
\frac{\partial q_\bth}{\partial \theta} + \frac{\partial}{\partial z}\left(q_\bth v\right)=0
\end{equation}

\vspace{-3pt}
\begin{figure}[t]
\begin{center}
\centerline{\includegraphics[width=\columnwidth]{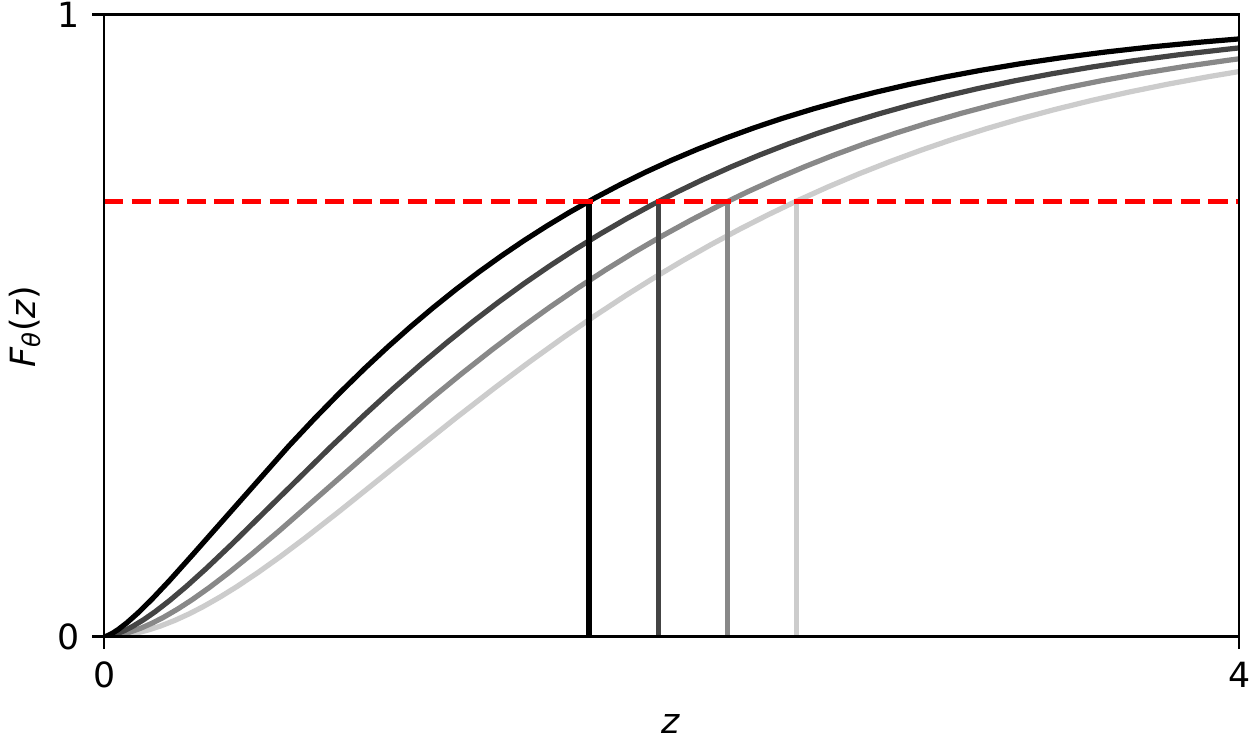}}
\caption{We illustrate how the pathwise derivative is obtained from the CDF in the univariate case.
  The black curves depict the CDF of the Gamma distribution with $\beta=1$ and $\alpha$ varying between 1.4 and 2.0. The
  red line corresponds to a fixed quantile $u$. As we vary $\alpha$ the point $z$ where the CDF intersects the red line varies.
  The rate of this variation is precisely the derivative  $\frac{d z}{d \alpha} $.}
\label{fig:regions}
\end{center}
\end{figure}

\subsubsection{Example: Truncated Unit Normal}
\label{sec:truncnormal}
We consider an illustrative case where Eqn.~\ref{eqn:mf} can be computed 
in closed form.
For simplicity we consider the unit Normal distribution truncated\footnote{As one 
would expect, Eqn.~\ref{eqn:mf}
yields the standard reparameterized gradient in the case of an non-truncated Normal distribution. Also note that the truncated
unit normal is amenable to the reparameterization trick provided that one can compute the inverse error function ${\rm erf}^{-1}$.}
to the interval  $[0, \kappa]$ with $\kappa$ as the only free parameter. A simple computation yields
\begin{equation}
\frac{dz}{d\kappa}  = e^{\tfrac{1}{2}(z^2-\kappa^2)} \frac{\rm{erf}(\tfrac{z}{\sqrt{2}})}{\rm{erf}(\tfrac{\kappa}{\sqrt{2}})}
\end{equation}
First, notice that for $z=\kappa$ we have $\frac{dz}{d\kappa}=1$, which is what we would expect, since $u = 1$ is 
mapped to the rightmost edge of the interval at $z = \kappa$, i.e. $F_\kappa^{-1}(1) = \kappa$. Similarly we have
$\frac{dz}{d\kappa}=0$ for $z=0$. For $z\in(0,\kappa)$ the derivative $\frac{dz}{d\kappa}$ interpolates smoothly between 0 and 1. This makes sense, since for a fixed value of $u$ as we get further into the tails of the distribution, nudging $\kappa$ to the right has a correspondingly larger effect on $z=F_\kappa^{-1}(u)$, while it has a correspondingly smaller effect for $u$ in the bulk of the distribution.

\subsubsection{Example: Univariate Mixture Distributions}

Consider a mixture of univariate distributions:
\begin{equation}
q_\bth(z) = \sum_{k=1}^{K} \pi_k q_{\theta_k}(z)
\end{equation}
If we have analytic control over the individual CDFs (or know how to approximate them and their derivatives w.r.t.~the parameters) then
we can immediately appeal to Eqn.~\ref{eqn:mf}. Concretely for derivatives w.r.t.~the parameters of each
component distribution we have:
\begin{equation}
\frac{\partial z}{\partial \theta_i} = - \frac{\pi_i \frac{\partial F_{\theta_i}}{\partial {\theta_i}}(z)}{q_\bth(z) } 
\end{equation}
from which we can get, for example
\begin{equation}
\frac{\partial z}{\partial \mu_i} = \frac{\pi_i q_{\mu_i, \sigma_i}(z) }{q_\bth(z) } 
\end{equation}
for a mixture of univariate Normal distributions.

\begin{figure}[t]
\begin{center}
\centerline{\includegraphics[width=.8\columnwidth]{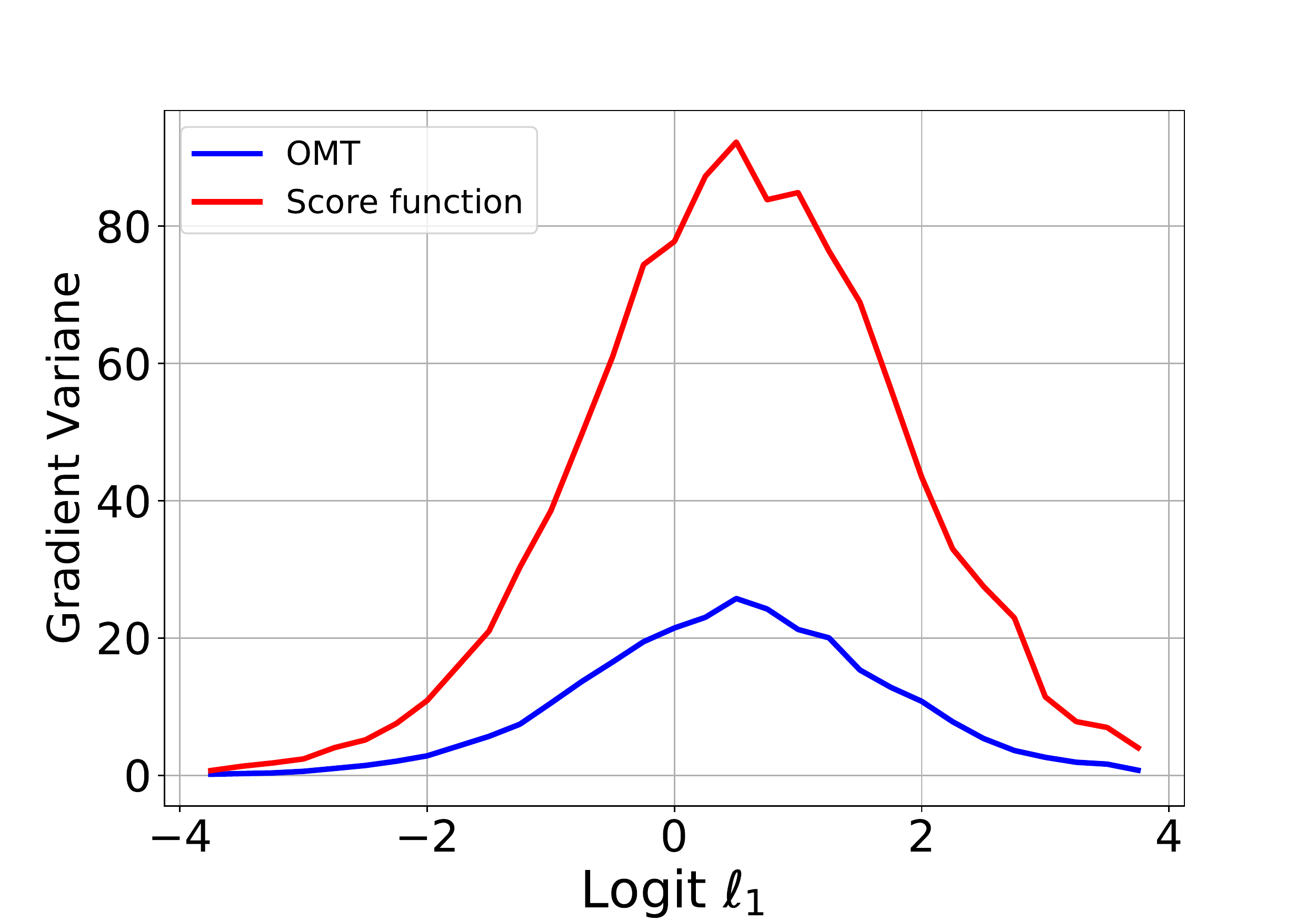}}
\caption{We compare the OMT  gradient to the score function gradient for the test function $f(z) = z^4$ where
$q_\bth(z)$ is a mixture with two components. Depicted is the variance of the gradient w.r.t.~the logit $\ell_1$ that governs
the mixture probability of the first component. The logit of the second component is fixed to be zero.}
\label{fig:mix}
\end{center}
\end{figure}

In Fig.~\ref{fig:mix} we demonstrate that the OMT gradient for a mixture of univariate Normal distributions
can have much lower variance than the corresponding score function gradient. Here the mixture has two components
with $\bm{\mu} = (0, 1)$ and $\bm{\sigma} = (1, 1)$. Note that using the reparameterization trick in this setting
would be impractical.

\vspace{-3pt}

\subsection{The Multivariate Case}

\newcommand*\VF[1]{\mathbf{#1}}
\newcommand*\dif{\mathop{}\!\mathrm{d}}

Suppose we are given a velocity field that satisfies the transport equation:
\begin{equation}
\label{eqn:transportsupp}
\frac{\partial}{\partial \theta} q_\bth + \nabla_{\bm z} \cdot \left( q_\bth {\bm v^\theta}\right)=0
\end{equation}
Then, as discussed in the main text, we can form the gradient estimator
\begin{equation}
\label{eqn:multiestsupp}
\nabla_\theta \mathcal{L} = \E_{ q_\bth(\bz) } \! \left[ \bm{v}^\theta  \cdot \nabla_{\bf z} f  \right]
\end{equation}
That this gradient estimator is unbiased follows directly from the transport equation and divergence theorem:
\begin{equation}
\begin{split}
&\nabla_\theta \mathcal{L}  = \int d\bz \frac{\partial q_\bth(\bz)}{\partial \theta} f(\bz) 
=-\! \int d\bz \nabla_\bz \cdot \left( q_\bth {\bm v}^\theta \right)f(\bz) 
= \\ &\int d\bz  q_\bth(\bz) \nabla_\bz  f \cdot {\bm v}^\theta 
=\E_{q_\bth(\bz)} \left[ \nabla_\bz f \cdot {\bm v}^\theta  \right]
\end{split}
\end{equation}
where we appeal to the identity
\begin{equation}
\begin{split}
\int_{V} f \nabla_\bz \! \cdot \! (q_\bth {\bm v}^\theta) \dif V = 
&-\int_{V} \nabla_\bz f \cdot (q_\bth {\bm v}^\theta) \dif V + \\ 
&\oint_{S} (q_\bth f {\bm v}^\theta) \cdot \hat{\VF{n}} \dif S
\end{split}
\end{equation}
and assume that $q_\bth f{\bm v}^\theta $ is sufficiently well-behaved that we can drop the surface integral.
This is just the multivariate generalization of the derivation in the previous section.

\subsection{Multivariate Normal}
\subsubsection{Whitened Coordinates}

First we take a look at gradient estimators in whitened coordinates $\bm{\tilde{z}} = L^{-1} \bm{z}$.
The reparameterization trick ansatz for the velocity field can be obtained by transforming the solution in Eqn.~\ref{eqn:rtvecsupp} (which is also
given in the main text) to the new coordinates:
\begin{equation}
\tilde{v}_i \equiv \frac{\partial \tilde{z}_i}{\partial L_{ab}}  = L^{-1}_{ia} \tilde{z}_b
\end{equation}
Note that the transport equation for the multivariate distribution can be written in the form
\begin{equation}
\label{eqn:mvnlogcont}
\frac{\partial}{\partial L_{ab}} \log{q} + \nabla \cdot {\bm{ \tilde{v}}} + {\bm{\tilde{v}}} \cdot \nabla \log q=0
\end{equation}
The homogenous equation (i.e.~the transport equation without the source term $\frac{\partial \log q }{\partial L_{ab}} $) 
is then given by
\begin{equation}
\nabla \cdot \bm{\tilde{v}} = \bm{\tilde{v}} \cdot \bm{\tilde{z}} 
\end{equation}
In these coordinates it is evident that infinitesimal rotations, i.e.~vector fields of the form
\begin{equation}
\tilde{w}_i = (A \tilde{\bm{z}})_i \qquad {\rm with} \qquad A_{ij} = -A_{ji}
\end{equation}
satisfy\footnote{These are in fact not the only solutions; in addition there are non-linear solutions.}
the homogenous equation,
since
\begin{equation}
\nabla \cdot \bm{\tilde{w}} = {\rm Tr} \;A = 0 = \sum_{ij}  \tilde{z}_i A_{ij} \tilde{z}_j = \bm{\tilde{w}} \cdot \bm{\tilde{z}} 
\end{equation}
Finally, if we make the specific choice
\begin{equation}
A_{ij} = \frac{1}{2} \left(  \delta_{ib} L^{-1}_{ja}  - \delta_{jb} L^{-1}_{ia} \right)
\end{equation}
we find that $\tilde{v}_i + \tilde{w}_i$ (which automatically satisfies the transport equation)
and which is given by
\begin{equation}
\label{eqn:vw}
\nonumber
\tilde{v}_i + \tilde{w}_i \equiv \left(\frac{\partial \tilde{z}_i}{\partial L_{ab}} \right)^{\rm OMT} \!\! = 
\frac{1}{2} \left( L^{-1}_{ia} \tilde{z}_b + \delta_{ib} \sum_k L^{-1}_{ka} \tilde{z}_k \right)
\end{equation}
satisfies the symmetry condition
\begin{equation}
\frac{\partial}{\partial \tilde{z}_j} \left(\frac{\partial \tilde{z}_i}{\partial L_{ab}} \right)^{\rm OMT} =
 \frac{\partial}{\partial \tilde{z}_i} \left(\frac{\partial \tilde{z}_j}{\partial L_{ab}} \right)^{\rm OMT} 
\end{equation}
since
\begin{equation}
\frac{\partial}{\partial \tilde{z}_j} \left(\frac{\partial \tilde{z}_i}{\partial L_{ab}} \right)^{\rm OMT} =
\frac{1}{2} \left( L^{-1}_{ia} \delta_{jb} + L^{-1}_{ja} \delta_{ib}   \right)
\end{equation}
which is symmetric in $i$ and $j$.
This implies that the velocity field can be specified as the gradient of a scalar field (this is generally true for the OMT solution), i.e.
\begin{equation}
\left(\frac{\partial \tilde{z}_i}{\partial L_{ab}} \right)^{\rm OMT}= \frac{\partial}{\partial \tilde{z}_i} \tilde{T}^{ab}(\bm{\tilde{z}})
 \end{equation}
for some $\tilde{T}^{ab}(\bm{\tilde{z}})$, which is evidently given by\footnote{Up to an unspecified additive constant.}
\begin{equation}
\tilde{T}^{ab}(\bm{\tilde{z}}) = \frac{1}{2} (L^{-{\rm T}} \tilde{z})_a \tilde{z}_b
 \end{equation}
Note, however, that this is not the OMT solution we care about: it minimizes a \emph{different} kinetic 
energy functional to the one we care about (namely it minimizes the kinetic energy functional in whitened coordinates and not in natural coordinates).

We now explicitly show that solutions of the transport equation that are modified by the addition
of an infinitesimal rotation (as in Eqn.~\ref{eqn:vw}) still yield valid gradient estimators. 
Consider a test statistic $f(\bm{\tilde{z}})$ that is a monomial in $\bm{\tilde{z}}$:
\begin{equation}
f(\bm{\tilde{z}}) = \kappa \prod_{i=1}^{n}  \tilde{z}_i ^ {n_i}
\end{equation}
It is enough to show that the following expectation vanishes:\footnote{Note that we can thus think of this term as a control variate.}
\begin{equation}
\label{eqn:mvngrad}
\E_{q_\bth(\bm{\tilde{z}})} \left[ \sum_{ij} \frac{\partial f }{\partial \tilde{z}_i} A_{ij} \tilde{z}_j \right]
\end{equation}
where $A_{ij}$ is an antisymmetric matrix. The sum in Eqn.~\ref{eqn:mvngrad} splits up into a sum of 
paired terms of the form
\begin{equation}
\label{eqn:mvngrad2}
\E_{q_\bth(\bm{\tilde{z}})} \left[  A_{ij} \left( \frac{\partial f }{\partial \tilde{z}_i}  \tilde{z}_j -  \frac{\partial f }{\partial \tilde{z}_j}  \tilde{z}_i\right) \right]
\end{equation}
We can easily show that each of these paired terms has zero expectation. First note that the expectation
is zero if either of $i$ or $j$ is even (since $\E_{q_\bth(\bm{\tilde{z}})} \left[ \tilde{z}_l^{2k-1} \right] = 0$). If both $i$ and $j$ are odd we get (using 
$\E_{q_\bth(\bm{\tilde{z}})} \left[ \tilde{z}_l^{2k} \right] = (2k-1)!!$, where $!!$ is the double factorial)
\begin{equation}
\label{eqn:mvngrad3}
\kappa A_{ij} \left[ n_i (n_i\!-\!2)!! n_j!! \!-\! n_j (n_j\!-\!2)!! n_i!! \right] = 0
\end{equation} 
Thus, solutions of the transport equation that are modified by the addition
of an infinitesimal rotation still yield the same gradient $\nabla_{L_{ab}} \E_{q_\bth(\bm{\tilde{z}})} \left[ f(\bm{\tilde{z}}) \right]$ in expectation.

\subsection{Natural Coordinates}

We first show that the velocity field $\bm{v}^{\rm RT}$ that follows from the reparameterization trick satisfies the transport
equation in the (given) coordinates $\bm{z}$, where we have
\begin{equation}
\label{eqn:rtvecsupp}
v_i^{\rm RT} \equiv \frac{\partial z_i}{\partial L_{ab}} =  \delta_{ia}  (L^{-1} \bm{z})_b 
\end{equation}
We have that 
\begin{equation}
\begin{split}
\frac{\partial \log q }{\partial L_{ab}} &= \frac{\partial}{\partial L_{ab}} \left( -\log \det L - \frac{1}{2} \bm{z}^{\rm T} \Sigma^{-1} \bm{z} \right) \\
&= -L^{-1}_{ba} + \left( \Sigma^{-1} \bm{z} \right)_a \left(L^{-1} \bm{z} \right)_b
\end{split}
\end{equation}
and 
\begin{equation}
\nabla \cdot \bm{v}^{\rm RT} = L^{-1}_{ba} 
\end{equation}
and
\begin{equation}
\nonumber
\bm{v}^{\rm RT} \cdot \nabla \log q = - \bm{v}^{\rm RT} \cdot \left(\Sigma^{-1} \bm{z} \right) = - \left( \Sigma^{-1} \bm{z} \right)_a \left(L^{-1} \bm{z} \right)_b
\end{equation}
Thus, the terms cancel term by term and the transport equation is satisfied.

What about the OMT gradient in the natural (given) coordinates $\bm{z}$?
To proceed we represent $\bm{v}$ as a linear vector field with symmetric and antisymmetric parts.
Imposing the OMT condition determines the antisymmetric part.
Imposing the transport equation determines the symmetric part. We find that 
\begin{equation}
\label{eqn:mvnomt}
 v_i^{\rm OMT} =
\frac{1}{2} \left(  \delta_{ia}  (L^{-1} \bm{z})_b + z_a L^{-1}_{bi}   \right) + (S^{ab} \bm{z})_i
\end{equation}
where $S^{ab}$ is the unique symmetric matrix that satisfies the equation
\begin{equation}
\nonumber
\label{eqn:S}
\Sigma^{-1} S^{ab} + S^{ab} \Sigma^{-1} = \Xi^{ab} \;\; {\rm with} \;\; \Xi^{ab} \equiv \xi^{ab} + (\xi^{ab})^{\rm T} 
 \end{equation}
where we define 
\begin{equation}
\xi_{ij}^{ab} = \tfrac{1}{2} \left(L^{-1}_{bi} \Sigma_{aj}^{-1} \!-\! \delta_{ai} (L^{-1}\Sigma^{-1})_{bj} \right)
 \end{equation}
To explicitly solve Eqn.~\ref{eqn:S} for $S^{ab}$ we use SVD to write
\begin{equation}
 \Sigma^{-1} = UDU^{\rm T} \qquad {\rm and} \qquad\tilde{\Xi}^{ab} = U^{\rm T} \Xi^{ab} U
\end{equation}
where $D$ and $U$ are diagonal and orthogonal matrices, respectively. Then we have that
\begin{equation}
 S^{ab} = U \left( \tilde{\Xi}^{ab} \div (D \otimes \mathbb{1} \!+\! \mathbb{1} \otimes D) \right) U^{\rm T}
\end{equation}
where $\div$ represents elementwise division and $\otimes$ is the outer product. Note that a naive
implementation of a gradient estimator based on Eqn.~\ref{eqn:mvnomt} would explicitly construct
$\xi_{ij}^{ab}$, which has size quartic in the dimension. A more efficient implementation will instead
make use of $\xi_{ij}^{ab}$'s structure as a sum of products and never explicitly constructs $\xi_{ij}^{ab}$.\footnote{Our implementation
can be found here: \\ {\tiny \texttt{https://github.com/uber/pyro/blob/0.2.1/pyro/distributions/omt\char`_mvn.py}}}

\subsubsection{Bivariate Normal distribution}

\begin{figure}[t]
\begin{center}
\centerline{\includegraphics[width=.7\columnwidth]{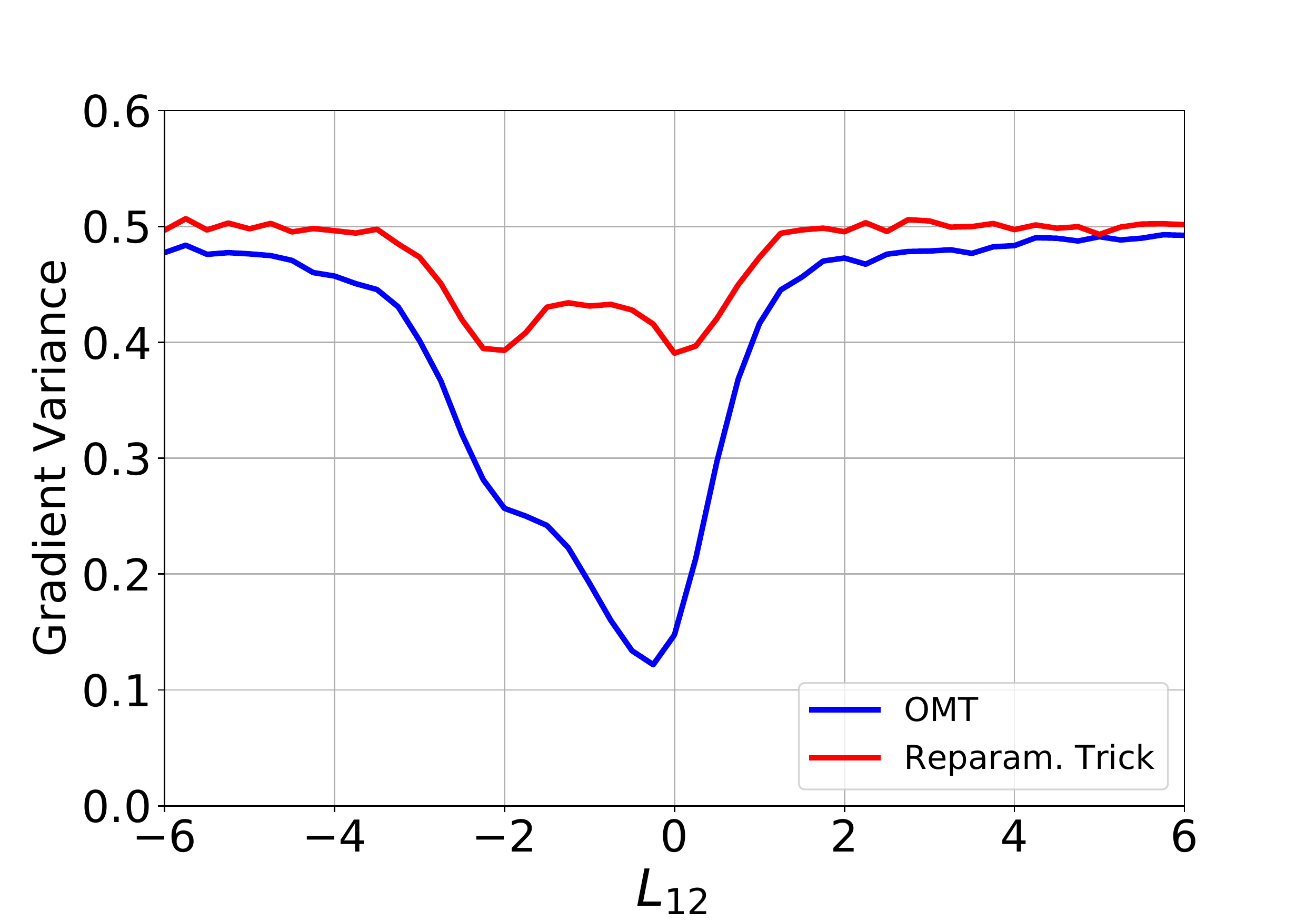}}
\caption{We compare the OMT gradient to the gradient from the reparameterization trick
for a bivariate Normal distribution and the test function $f_\bth(\bz) \!= \cos \bm{\omega} \cdot \bm{z}$ with $\bm{\omega}=(1,1)$. The Cholesky factor $\bm{L}$ has diagonal
elements $(1, 1)$ and off-diagonal element $L_{21}$. The gradient is with respect to $L_{21}$. The
variance for the OMT gradient is everywhere lower than for the reparameterization trick
gradient.}
\label{fig:bvn}
\end{center}
\end{figure}

In Fig.~\ref{fig:bvn} we compare the performance of our OMT gradient for a bivariate Normal distribution
to the reparameterization trick gradient estimator. 
We use a test function $f_\bth(\bz)$ for which we can compute the gradient exactly. We see that
the OMT gradient estimator performs favorably over the entire range of
parameters considered.

\vspace{-3pt}

\subsection{Gradient Variance for Linear Test Functions}

We  use the following example to give more intuition for when we expect OMT gradients for the multivariate Normal distribution to be lower variance
than RT gradients.
Let $q_\bth(\bz)$ be the unit normal distribution in $D$ dimensions. Consider the test function
\begin{equation}
f(\bz) = \sum_{i=1}^D \kappa_i z_i \qquad \mathcal{L} = \E_{q_\bth(\bz)} \left[ f(\bz) \right]
\end{equation}
and the derivative w.r.t.~the off-diagonal elements of the Cholesky factor $L$. A simple computation yields the total variance of the RT estimator:
\begin{equation}
\sum_{a > b} \rm{Var}\left(\frac{\partial \mathcal{L}}{\partial L_{ab}}\right) = \sum_{a > b} \kappa_a^2
\end{equation}
Similarly for the OMT estimator we find
\begin{equation}
\sum_{a > b} \rm{Var}\left(\frac{\partial \mathcal{L}}{\partial L_{ab}}\right) = \frac{1}{4} \sum_{a > b} \left( \kappa_a^2 + \kappa_b^2 \right)
\end{equation}
So if we draw the parameters $\kappa_i$ from a generic prior we expect the variance of the OMT estimator to be about half
of that of the RT estimator. Concretely, if $\kappa_i \sim \mathcal{N}(0, 1)$ then the variance of the OMT estimator will be exactly
half that of the RT estimator in expectation. While this computation is for a very specific case---a linear test function and a unit normal
$q_\bth(\bz)$---we find that this magnitude of variance reduction is typical.

\subsection{The Lugannani-Rice Approximation}

Saddlepoint approximation methods take advantage
of cumulant generating functions (CGFs) to construct (often very accurate) approximations to
probability density functions in situations where full analytic control is intractable.\footnote{We
refer the reader to ~\cite{butler2007saddlepoint} for an overview.} These methods are
also directly applicable to CDFs, where a particularly useful approximation---often used by
statisticians to estimate various tail probabilities---has been developed
by Lugannani and Rice \cite{lugannani1980saddle}. This approximation---after additional differentiation
w.r.t.~the parameters of the distribution $q_\bth(z)$---forms the basis of our approximate formulas for pathwise gradients
for the Gamma, Beta and Dirichlet distributions in regions of $(z, \theta)$ where the (marginal) density is approximately gaussian. As we
will see these approximations attain high accuracy.

For completeness we briefly describe the Lugannani-Rice approximation.
It is given by:
\begin{equation}
\label{eqn:lr}
    F(z) \approx 
\begin{cases}
    \Phi(\hat{w}) +  \phi(\hat{w})(1/\hat{w} - 1/\hat{u}) &\text{if } z \ne \mu\\
    \frac{1}{2} + \frac{K^{\prime\prime\prime}(0)}{6 \sqrt{2 \pi}K^{\prime\prime}(0)^{3/2}}              & \text{if } z= \mu\\
\end{cases}
\end{equation}
where 
\begin{equation}
\hat{w} = {\rm sgn}(\hat{s})\sqrt{2\{\hat{s}z - K(\hat{s})\}} \qquad  \hat{u} = \hat{s}\sqrt{K^{\prime\prime}(\hat{s})}
\end{equation}
and where $\hat{w} $ and $\hat{u}$ are functions of $z$ and the saddlepoint $\hat{s}$, with the saddlepoint defined implicitly by the equation
$K^\prime(\hat{s})=z$.  Here $K(s) = \log \mathbb{E}_{q_{\bth}(z)} \left[ \exp(sz)\right]$ is the CGF of $q_\bth(z)$, $\mu$ is the mean of $q_\bth(z)$, and
$\Phi(\cdot)$ and $\phi(\cdot)$ are the CDFs and probability densities of the unit normal distribution.
Note that Eqn.~\ref{eqn:lr} appears to have a singularity at $z=\mu$;  it can be shown, however,  that Eqn.~\ref{eqn:lr} is in fact smooth
at $z=\mu$. Nevertheless, in our numerical recipes we will need to take care to avoid numerical instabilities
near $z=\mu$ that result from finite numerical precision.

\subsection{Gamma Distribution}

Our numerical recipe for $\frac{dz}{d\alpha}$ for the standard Gamma distribution with $\beta=1$
divides $(z, \alpha$) space into three regions. If $z<0.8$ we use the Taylor series expansion given in the main text.
If $\alpha > 8$ we use the following set of expressions derived from the Lugannani-Rice approximation.
Away from the singularity, for $z \gtrless \alpha \pm \delta \cdot \alpha$, we use:
\begin{equation}
\label{eqn:gammaricefar}
\frac{dz}{d\alpha} =
\frac{
 \sqrt{\frac{2}{\alpha}}\frac{\alpha + z}{(\alpha-z)^2} \!+\! \log \frac{z}{\alpha} \! \left( \frac{\sqrt{8 \alpha}}{z \!-\! \alpha} \pm (z \!-\! \alpha \!-\! \alpha \log \tfrac{z}{\alpha})^{-\tfrac{3}{2}} \right)}
{\sqrt{8\alpha}/(z \mathcal{S}_{\alpha})}
\end{equation}
where
\begin{equation}
\label{eqn:stirling}
\nonumber
\mathcal{S}_{\alpha} \equiv 1 \!+\!\tfrac{1}{12 \alpha}\!+\!\tfrac{1}{288 \alpha^2}
\end{equation}
Near the singularity, i.e.~for $|z-\alpha| \le \delta \cdot \alpha$, we use:
\begin{equation}
\label{eqn:gammaricenear}
\frac{dz}{d\alpha} \!=\!
\frac{1440 \alpha^3 \!+\! 6\alpha z (53\!-\!120 z) \!-\! 65 z^2 \!+\! \alpha^2 (107 \!+\! 3600 z)}{1244160 \alpha^5 / (1 + 24 \alpha + 288 \alpha^2)}
\end{equation}
Note that Eqn.~\ref{eqn:gammaricenear} is derived from Eqn.~\ref{eqn:gammaricefar} by a Taylor expansion in powers of $(z-\alpha)$. 
We set $\delta = 0.1$, which is chosen to balance use of Eqn.~\ref{eqn:gammaricefar} (which is more accurate)
and Eqn.~\ref{eqn:gammaricenear} (which is more numerically stable for $z \approx \alpha$). Finally, in the remaining
region ($z>0.8$ and $\alpha<8$) we use a bivariate rational polynomial approximation $f(z,\alpha) = \exp\left(\frac{p(z,\alpha)}{q(z,\alpha)}\right)$ where $p,q$ are polynomials in the coordinates
$\log(z/\alpha)$ and $\log(\alpha)$,
with terms up to order 2 in $\log(z/\alpha)$ and order 3 in $\log(\alpha)$.
We fit the rational approximation using least squares on 15696 random $(z,\alpha)$ pairs
with $\alpha$ sampled log uniformly between 0.00001 and 10, and $z$ sampled conditioned on $\alpha$.
Our complete approximation for $\frac{dz}{d\alpha}$ is unit tested to have relative accuracy of 0.0005 on a wide range of inputs.

\subsection{Beta Distribution}

The CDF of the Beta distribution is given by 
\begin{equation}
\label{eqn:betacdf}
F_{\alpha, \beta}(z) = \frac{B(z; \alpha, \beta)}{B(\alpha, \beta)}
\end{equation}
where $B(z; \alpha, \beta)$ and $B(\alpha, \beta)$ are the incomplete beta function and beta function, respectively.
Our numerical recipe for computing $\frac{dz}{d\alpha}$ and $\frac{dz}{d\beta}$ for the Beta distribution
divides $(z, \alpha, \beta$) space into three sets of regions. First suppose that $z\ll1$. Then just like for the Gamma
distribution, we can compute a Taylor series of $B(z; \alpha, \beta)$ in powers of $z$
\begin{equation}
\label{eqn:betataylor}
B(z; \alpha, \beta) = z^\alpha \left( \frac{1}{\alpha} + \frac{1 - \beta}{1+\alpha}z + \frac{1-\tfrac{3\beta}{2}+\tfrac{\beta^2}{2}}{2 + \alpha} z^2 + ...\right)
\end{equation}
that can readily be differentiated w.r.t.~either $\alpha$ or $\beta$. Combined with the derivatives of the beta function, 
\begin{equation}
\label{eqn:betaderiv}
\begin{split}
\frac{d}{d\alpha}B(\alpha, \beta) &= B(\alpha, \beta) \left( \psi(\alpha) - \psi(\alpha+\beta)\right) \\
\frac{d}{d\beta}B(\alpha, \beta) &= B(\alpha, \beta) \left( \psi(\beta) - \psi(\alpha+\beta)\right)
\end{split}
\end{equation}
this gives a complete recipe for approximating $\frac{dz}{d\alpha}$ and $\frac{dz}{d\beta}$ for small $z$.\footnote{Here $\psi(\cdot)$ is the digamma function, which is available in most advanced tensor libraries.} By appealing to the symmetry of the Beta distribution
\vspace{-2pt}
\begin{equation}
\label{eqn:betasymm}
\rm{Beta}(z| \alpha, \beta) = \rm{Beta}(1-z| \beta, \alpha)
\end{equation}
we immediately gain approximations to $\frac{dz}{d\alpha}$ and $\frac{dz}{d\beta}$ for $1-z \ll1$. It remains to
specify when these various approximations are applicable. Let us define $\xi = z (1-z) (\alpha + \beta)$. Empirically
we find that these approximations are accurate for $\frac{dz}{d\alpha}$ if 
\begin{enumerate}[topsep=0pt,itemsep=0ex,partopsep=1ex,parsep=1ex]
\item $z\le0.5$ and $\xi<2.5$; or
\item $z\ge0.5$ and $\xi<0.75$
\end{enumerate}
with the conditions flipped for $\frac{dz}{d\beta}$. Depending on the precise region, we use 8 to 10 terms in the
Taylor series.

Next we describe the set of approximations we derived from the Lugannani-Rice approximation and that we find
 to be accurate for $\alpha > 6$ and $\beta > 6$. By Eqn.~\ref{eqn:betasymm} it is sufficient to describe
our approximation for $\frac{dz}{d\alpha}$.
First define $\sigma = \frac{\sqrt{\alpha \beta}}{(\alpha+\beta)\sqrt{\alpha+\beta+1}}$, the standard deviation of the Beta distribution.
Then away from the singularity, for $z \gtrless  \frac{\alpha}{\alpha+\beta} \pm \epsilon \cdot \sigma$, we use:
\begin{equation}
\label{eqn:betaricefar}
\frac{dz}{d\alpha} =
\frac{z (1-z)\left(\mathcal{A} + \log \frac{\alpha}{z (\alpha+\beta)} \mathcal{B}_{\pm}\right)}
{{\sqrt{\frac{2 \alpha \beta}{\alpha + \beta}}}\frac{S_{\alpha\beta}}{S_{\alpha}S_{\beta}}}
\end{equation}
with
\begin{equation}
\nonumber
\mathcal{A} = \frac{\beta(2\alpha^2(1-z)+\alpha \beta(1-z) +\beta^2z)}{\sqrt{2\alpha \beta}(\alpha+\beta)^{3/2}(\alpha(1-z)-\beta z)^2}
\end{equation}
and
\begin{equation}
\nonumber
\mathcal{B}_{\pm} =\frac{\sqrt{\frac{2 \alpha \beta}{\alpha + \beta}}}{\alpha(1\!-\!z)\!-\!\beta z} \pm 
\frac{1}{2}\left(\!\alpha \log \tfrac{\alpha}{(\alpha \!+\! \beta)(1\!-\! z)} \!+\! \beta \log \tfrac{\beta}{(\alpha\!+\! \beta)z}\right)^{-3/2}
\end{equation}
Near the singularity, i.e.~for $|z- \frac{\alpha}{\alpha+\beta}| \le \epsilon \cdot \sigma$, we use:
\begin{equation}
\label{eqn:betaricenear}
\frac{dz}{d\alpha} \!=\!  \frac{(12\alpha+1)(12\beta+1)(\mathcal{H} + \mathcal{I}  + \mathcal{J}  +\mathcal{K}  )}{12960 \alpha^3 \beta^2 (\alpha+\beta)^2(12\alpha+12\beta+1)}
\end{equation}
with 
\begin{equation}
\nonumber
\begin{split}
\mathcal{H}  &\!=\!   8\alpha^4(135\beta-11)(1-z)   \\
\mathcal{I}  &\!=\!   \alpha^3\beta(453-455z+1620\beta(1-z))  \\
\mathcal{J}  &\!=\!     3\alpha^2\beta^2(180\beta-90z+59) \\
\mathcal{K}  &\!=\!    \alpha\beta^3(20z(27\beta + 16)+43)  +47\beta^4z \\
\end{split}
\end{equation}
We set $\epsilon = 0.1$, which is chosen to balance numerical accuracy and numerical stability (just
as in the case of the Gamma distribution).  

Finally, in the remaining region we use a rational multivariate polynomial
approximation
\begin{equation}
\nonumber
  f(z, \alpha, \beta) = \frac {p(z,\alpha,\beta)} {q(z,\alpha,\beta)}
                        \frac{z (1 - z)}{\beta} \left(\psi(\alpha + \beta) - \psi(\alpha)\right)
\end{equation}
where $p,q$ are polynomials in the three coordinates
$\log(z)$, $\log(\alpha / z)$, and $\log((\alpha + \beta)z/\alpha)$
with terms up to order 2, 2, and 3 in the respective coordinates.
The rational approximation was minimax fit to 2842 points in the
remaining region for  $0.01 < \alpha,\beta < 1000$.
Test points were randomly sampled using log uniform sampling of $\alpha,\beta$
and stratified sampling of $z$ conditioned on $\alpha,\beta$.
Minimax fitting achieved about half the maximum error of simple least squares fitting.
Our complete approximation for $\frac{dz}{d\alpha}$ and $\frac{dz}{d\beta}$ is unit tested to have relative accuracy of 0.001 on a wide range of inputs.

\vspace{-3pt}
\begin{figure}[t]
\begin{center}
\centerline{\includegraphics[width=\columnwidth]{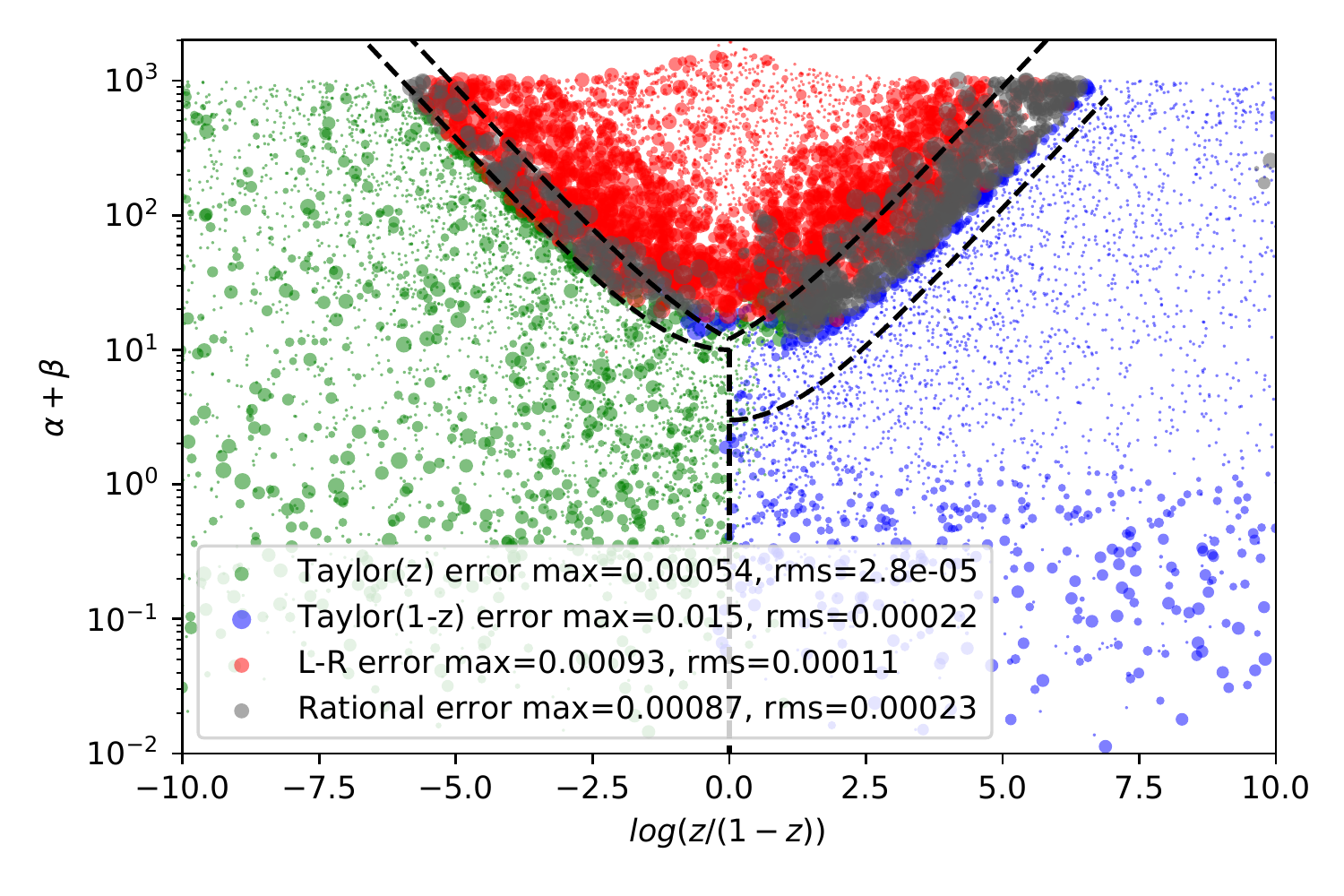}}
\caption{Relative error of our four approximations for $\frac{dz}{d\alpha}$ for the Beta distribution in their respective regions.
  Note that the region boundaries are in the three-dimensional $z,\alpha,\beta$ space,
  so the upper boundaries are only cross-sections.}
\label{fig:regions}
\end{center}
\end{figure}

\subsection{Dirichlet Distribution}

For completeness we record the general version of the formula for the pathwise gradient (given implicitly in the main text):
\begin{equation}
\label{eqn:dirichletmaster2}
\frac{dz_i}{d\alpha_j} = -\frac{\frac{\partial F_{{\rm Beta}}}{\partial \alpha_j }(z_j | \alpha_j, \alpha_{\rm tot} - \alpha_j)}{{\rm Beta}(z_j | \alpha_j, \alpha_{\rm tot} - \alpha_j)} \times \left(  \frac{\delta_{ij}  - z_i}{1-z_j} \right)
\end{equation}
We want to confirm that Eqn.~\ref{eqn:dirichletmaster2} satisfies the transport equation for each choice of $j=1,...,n$:
\begin{equation}
\label{eqn:logcont}
\frac{\partial}{\partial \alpha_j} \log{q} + \nabla \cdot {\bm{v}} + \bm{v} \cdot \nabla \log q=0
\end{equation}
Treating $z_j$ as a function of ${\bf z}_{-j}=(z_1, ..., z_{j-1}, z_{j+1}, ..., z_n)$ everywhere and introducing
obvious shorthand for $F_{\rm Beta}(\cdot)$ and $\rm{Beta}(\cdot)$ we have:
\begin{equation}
\nonumber
\begin{split}
\nabla \cdot \bm{v} &= \sum_{i \ne j} \frac{\partial}{\partial z_i} \left(  \frac{\frac{\partial F_{{\rm Beta}}}{\partial \alpha_j }(z_j | \alpha_j, \alpha_{\rm tot} - \alpha_j)}
{{\rm Beta}(z_j | \alpha_j, \alpha_{\rm tot} - \alpha_j)} \frac{z_i}{\sum_{k \ne j} z_k}\right) \\
 &= \frac{\frac{\partial F}{\partial \alpha_j }} B  \frac{n-2}{1-z_j}  
- \frac{\partial \log B}{\partial \alpha_j} +\frac{\partial F}{\partial \alpha_j} \frac{ \left( \log B \right)^{\prime}  }{B}
\end{split}
\end{equation}
where $\left( \log B \right)^{\prime}$ is differentiated w.r.t.~the argument of $B(z_j)$. We further have that
\begin{equation}
\nonumber
\bm{v} \cdot \nabla \log q = \frac{\frac{\partial F}{\partial \alpha_j }} B \left( \sum_{i \ne j} \frac{\alpha_i - 1 }{1-z_j} - \frac{\alpha_j -1}{z_j}\right)  
\end{equation}
and
\begin{equation}
\nonumber
\frac{\partial}{\partial \alpha_j} \log{q}  = \psi(\alpha_j) - \psi(\alpha_{\rm tot}) + \log z_j
\end{equation}
Since we have
\begin{equation}
\nonumber
\frac{\partial \log B}{\partial \alpha_j}  = \psi(\alpha_j) - \psi(\alpha_{\rm tot}) + \log z_j
\end{equation}
and
\begin{equation}
\nonumber
\left( \log B \right)^{\prime} = \frac{\alpha_j -1}{z_j} - \frac{\alpha_{\rm tot} - \alpha_j - 1}{1-z_j}
\end{equation}
it becomes clear by comparing the individual terms that everything cancels identically and so Eqn.~\ref{eqn:logcont}
is in fact satisfied by the velocity field in Eqn.~\ref{eqn:dirichletmaster2}.

Finally, we note that Eqn.~\ref{eqn:dirichletmaster2} is \emph{not} the OMT solution in the coordinates $\bm{z}_{-j}$. It
\emph{is} the OMT solution in some coordinate system, but it is not readily apparent which coordinate system that might be.

\subsection{Student's t-Distribution}
As another example of how to compute pathwise gradients consider Student's t-distribution. Although we have not done so ourselves, it
should be straightforward to compute an accurate approximation to Eqn.~\ref{eqn:mf}. In the absence
of such an approximation, however, we can still get a pathwise gradient for the Student's t-distribution by
composing the Normal and Gamma distributions:
\begin{equation}
\label{eqn:student}
\begin{split}
&\tau \sim \rm{Gamma}(\nu/2,1) \qquad x | \tau \sim \mathcal{N}(0, \tau^{-\tfrac{1}{2}})   \\
& \Rightarrow z \equiv \sqrt{\tfrac{\nu}{2}} x  \sim \rm{Student}(\nu) 
\end{split}
\end{equation}
Since sampling $z$ like this introduces an auxiliary random degree of freedom, pathwise gradients 
$\frac{dz}{d\nu}$ computed using Eqn.~\ref{eqn:student} 
will exhibit a larger variance than a direct computation of Eqn.~\ref{eqn:mf} would 
yield.\footnote{Note, however, that this additional variance will decrease as $\nu$ increases.} 
The point is that
\emph{no additional work} is needed to obtain this particular form of the pathwise gradient: just use pathwise
gradients for the Gamma and Normal distributions and the sampling procedure in Eqn.~\ref{eqn:student}.

\subsection{Baseball Experiment}
\label{sec:baseball}

To gain more insight into when we expect the OMT gradient estimator for the multivariate Normal distribution to outperform the RT gradient estimator,
we conduct an additional experiment.
We consider a model for repeated binary trial data (baseball players at bat) using the data in \cite{efron1975data} and the modeling setup in \cite{stanmanual} with partial pooling. 
There are 18 baseball players and the data consists of 45 hits/misses for each player. 
The model has two global latent variables and 18 local latent variables so that the posterior is 20-dimensional. 
Specifically, the two global 
latent random variables are $\phi$ and $\kappa$, with priors
$\rm{Uniform}(0,1)$ and $\rm{Pareto}(1, 1.5) \propto \kappa^{-5/2}$, respectively. The local latent random variables are given by $\theta_i$ for $i=0,...,17$, with
$p(\theta_i) = \rm{Beta}(\theta_i | \alpha= \phi  \kappa, \beta=(1 - \phi)  \kappa)$. The data likelihood factorizes into 45 Bernoulli observations with mean chance of success $\theta_i$ for each player $i$. The variational approximation is formed in the unconstrained space
 $\{\rm{logit}(\phi), \log(\kappa - 1), \rm{logit}(\theta_i)\}$ and consists of a multivariate Normal distribution 
 with a full-rank Cholesky factor $\bm{L}$.
  We use the Adam optimizer for training with a learning rate of $5 \times 10^{-3}$ \cite{kingma2014adam}. 
  
\begin{figure}[t]
\begin{center}
\centerline{\includegraphics[width=\columnwidth]{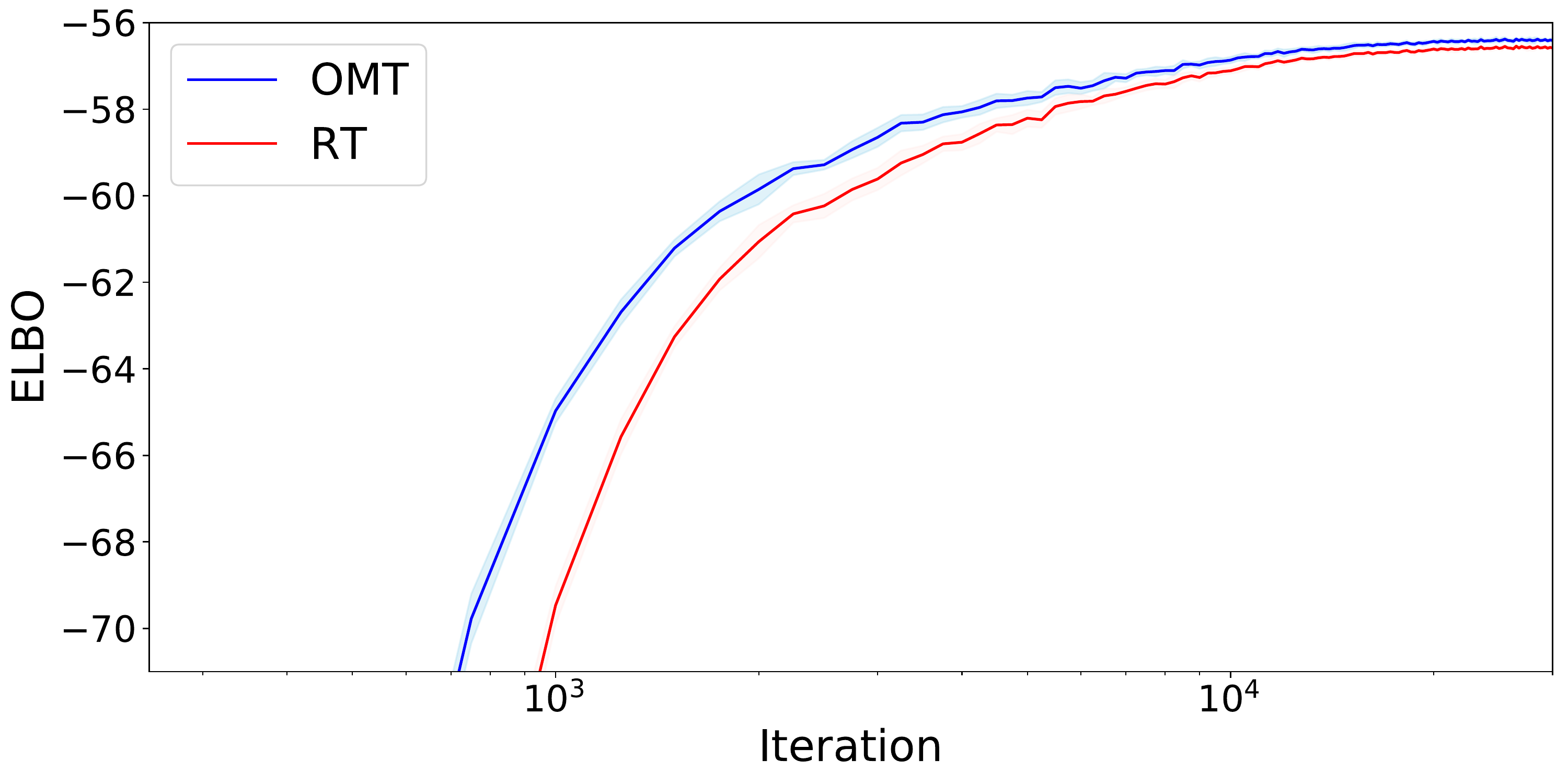}}
\caption{ELBO training curves for the experiment in Sec.~\ref{sec:baseball} for the case where the Cholesky factor
is initialized far from the identity. Depicted is the mean ELBO for 10 runs with $1\!-\!\sigma$ uncertainty bands 
around the mean. The OMT gradient estimator learns more quickly than the RT estimator and attains a higher ELBO.}
\label{fig:baseball}
\end{center}
\end{figure}

 For this particular model mean field SGVI performs reasonably well, since
 correlations between the latent random variables are not particularly strong. If we initialize $\bm{L}$ near the identity, we find that the
 OMT and RT gradient estimators perform nearly identically, with the difference that the former has an increased computational cost of about 25\% per iteration.
 If, however, we initialize $\bm{L}$ far from the identity---so that the optimizer has to traverse a considerable distance in $\bm{L}$ space where the covariance
 matrix exhibits strong correlations---we find that the OMT estimator makes progress more quickly than the RT estimator and converges to a higher ELBO,
 see Fig.~\ref{fig:baseball}. Generalizing from this, we expect the OMT gradient estimator for the multivariate Normal distribution
  to exhibit better sample efficiency than the RT estimator in problems where the covariance matrix exhibits strong correlations.
This is indeed the case for the GP experiment in the main text, where the learned kernel induces strong temporal correlations.

\vspace{-3pt}

\subsection{Experimental Details}

As noted in the main text, we use single-sample gradient estimators in all experiments.
Unless noted otherwise, we always include the score function term for \acrshort{RSVI}.

\subsubsection{Multivariate Normal Synthetic Test Function Experiment }

We describe the setup for the experiment corresponding to Fig.~5 in the main text.
The dimension is fixed to $D=50$ and the mean of $q_\bth$ 
is fixed to the zero vector. The Cholesky factor $\bm{L}$ that
enters into $q_\bth$ is constructed as follows. The diagonal of $\bm{L}$ consists of all ones. To construct the off-diagonal
terms we proceed as follows. We populate the entries below the diagonal of a matrix $\Delta \bm{L}$ by drawing each
entry from the uniform distribution on the unit interval. Then we define $\bm{L} = \mathbb{1}_D + r \Delta \bm{L}$. Here $r$
controls the magnitude of off-diagonal terms of $\bm{L}$ and appears on the horizontal axis of Fig.~5 in the main text.
The three test functions are constructed as follows. First we construct a strictly lower diagonal matrix $ \bm{Q}^\prime$ by
drawing each entry from a bernoulli distribution with probability 0.5. We then define $ \bm{Q}= \bm{Q}^\prime + \bm{Q}^{\prime T}$.
The cosine test function is then given by
\begin{equation}
f(\bz) = \cos \left(\sum_{i, j} Q_{ij} z_i / D \right)
\end{equation}
The quadratic test function is given by
\begin{equation}
f(\bz) = \bz^{T} \bm{Q} \bz
\end{equation}
The quartic test function is given by
\begin{equation}
f(\bz) = \left(\bz^{T} \bm{Q}\bz\right)^2 
\end{equation}
In all cases the gradients can be computed analytically, which makes it easier to reliably estimate the variance of the gradient
estimators.

\subsubsection{Sparse Gamma \acrshort{DEF}}

Following \cite{naesseth2017reparameterization}, we use analytic expressions for each entropy term (as opposed to using the sampling estimate).
We use the adaptive step sequence $\rho^n$ proposed by \cite{kucukelbir2016automatic} and
also used in \cite{naesseth2017reparameterization}, which combines \textsc{rmsprop} \cite{Tieleman2012} and Adagrad \cite{duchi2011adaptive}: 
\begin{equation}
\label{eqn:opt}
\begin{split}
\rho^n &= \eta \cdot n^{-1/2 + \delta} \cdot \left(1 + \sqrt{s^n}\right)^{-1}. \\
s^n &= t \left( \hat{g}^n \right)^2 + (1-t) s^{n-1}
\end{split}
\end{equation}
Here $n=1,2,...$ is the iteration number and the operations in Eqn.~\ref{eqn:opt} are to be understood element-wise. In our case
the gradient $\hat{g}^n$ is always a single-sample estimate.
We fix $\delta = 10^{-16}$ and $t=0.1$. In contrast to  \cite{kucukelbir2016automatic} but in line with \cite{naesseth2017reparameterization}
we initialize $s_0$ at zero. To choose $\eta$ we did a grid search for each gradient estimator and each of the two model variants. Specifically, for each $\eta$ we
did 100 training iterations for three trials with different random seeds and then chose the $\eta$  that yielded the highest mean ELBO after 100 iterations. 
This procedure led to the selection of $\eta=4.5$ for the first model variant and $\eta=30$ for the second model variant (note that within each model
variant the gradient estimators preferred the same value of $\eta$). For the first model variant we included the score function-like term in the \acrshort{RSVI}
gradient estimator,
while we did not include it for the second model variant, as we found that this hurt performance. In both cases we used the shape augmentation setting $B=4$, which was also used for the results reported in \cite{naesseth2017reparameterization}.
After fixing $\eta$ we trained the model for 2000 iterations, initializing with another random number seed. The figure in the main text shows the training curves for that single run.
We confirmed that other random number seeds give similar results. A reference implementation can be found here:
{\tiny \texttt{https://github.com/uber/pyro/blob/0.2.1/examples/sparse\char`_gamma\char`_def.py}}

\subsubsection{Gaussian Process Regression}

We used the Adam optimizer \cite{kingma2014adam} to optimize the ELBO with single-sample gradient estimates. We chose
the Adam hyperparameters by doing a grid search over the learning rate and $\beta_1$. For each combination $({\rm lr}, \beta_1)$ we
did 20 training iterations for three trials with different random seeds and then chose the combination that yielded the highest mean ELBO after 20 iterations. 
This procedure led to the selection of a learning rate of $0.030$ and $\beta_1 = 0.50$ for both gradient estimators (OMT and reparameterization trick).
We then trained the model for 500 iterations, initializing with another random number seed. The figure in the main text shows the training curves for that single run.
We confirmed that other random number seeds give similar results.

\end{document}